\documentclass[10pt,twocolumn,letterpaper]{article}

\usepackage{cvpr}
\usepackage{times}
\usepackage{epsfig}
\usepackage{graphicx}
\usepackage{amsmath}
\usepackage{amssymb}
\usepackage{color}
\usepackage{booktabs}

\newcommand{\quotes}[1]{``#1''}

\DeclareMathOperator*{\argmin}{argmin}

\usepackage[pagebackref=true,breaklinks=true,colorlinks,bookmarks=false]{hyperref}

\cvprfinalcopy 

\def\cvprPaperID{****} 

\ifcvprfinal\fi
\begin{document}

\title{A Photometrically Calibrated Benchmark For Monocular Visual Odometry}

\author{Jakob Engel and Vladyslav Usenko and Daniel Cremers\\
Technical University Munich
}

\maketitle

\begin{abstract}
We present a dataset for evaluating the tracking accuracy of 
monocular visual odometry and SLAM methods. It contains 50
real-world sequences comprising more than 100 minutes of video, 
recorded across dozens of different environments -- ranging from narrow 
indoor corridors to wide outdoor scenes. 
All sequences contain mostly exploring camera motion, starting and ending at the same position.
This allows to evaluate tracking accuracy via the accumulated drift from start 
to end, without requiring ground truth for the full sequence. 
In contrast to existing datasets, all sequences are photometrically calibrated.
We provide exposure times for each frame as reported by the sensor, 
the camera response function, and dense lens attenuation factors. 
We also propose a novel, simple approach to non-parametric vignette calibration, 
which requires minimal set-up and is easy to reproduce.
Finally, we thoroughly evaluate two existing methods (ORB-SLAM \cite{mur2015orb} and DSO \cite{engel16archiveOdometry}) on
the dataset, including an analysis of the effect of image resolution,
camera field of view, and the camera motion direction.
\end{abstract}

\section{Introduction}
Structure from Motion or Simultaneous Localization and Mapping (SLAM) has become an increasingly important topic, since
it is a fundamental building block for many emerging technologies -- from autonomous
cars and quadrocopters to virtual and augmented reality.
In all these cases, sensors and cameras built into the hardware are designed to produce
data well-suited for computer vision algorithms, instead of capturing images optimized for human viewing.
In this paper we present a new monocular visual odometry (VO) / SLAM evaluation benchmark, that attempts to
resolve two current issues:

\begin{figure}
\centering
\includegraphics[width=.14\linewidth]{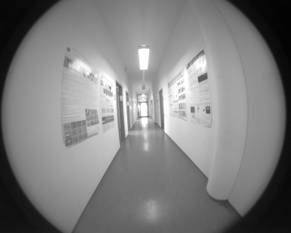}\hspace{-0.7mm}
\includegraphics[width=.14\linewidth]{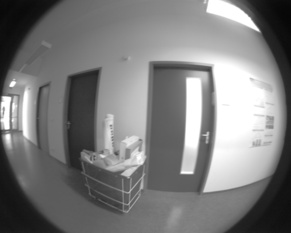}\hspace{-0.7mm}
\includegraphics[width=.14\linewidth]{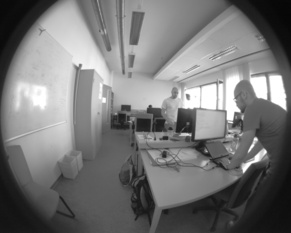}\hspace{-0.7mm}
\includegraphics[width=.14\linewidth]{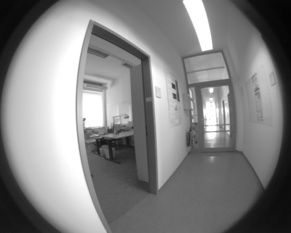}\hspace{-0.7mm}
\includegraphics[width=.14\linewidth]{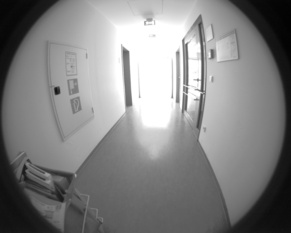}\hspace{-0.7mm}
\includegraphics[width=.14\linewidth]{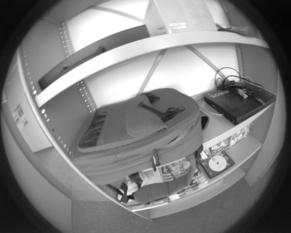}\hspace{-0.7mm}
\includegraphics[width=.14\linewidth]{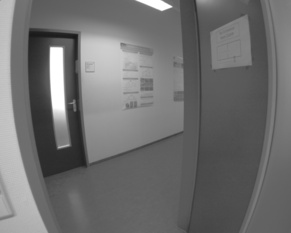}\hspace{-0.7mm}\\
\includegraphics[width=.14\linewidth]{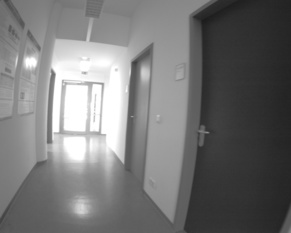}\hspace{-0.7mm}
\includegraphics[width=.14\linewidth]{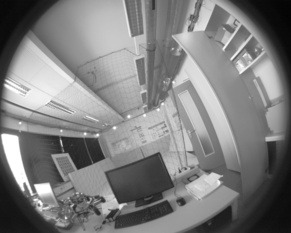}\hspace{-0.7mm}
\includegraphics[width=.14\linewidth]{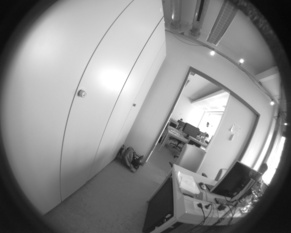}\hspace{-0.7mm}
\includegraphics[width=.14\linewidth]{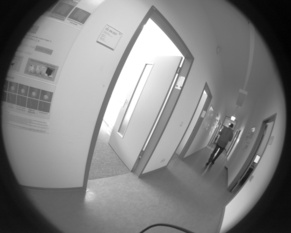}\hspace{-0.7mm}
\includegraphics[width=.14\linewidth]{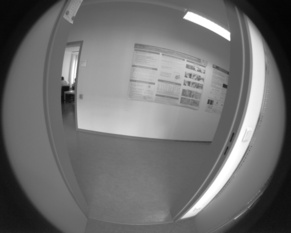}\hspace{-0.7mm}
\includegraphics[width=.14\linewidth]{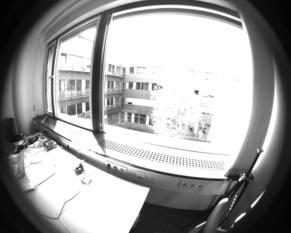}\hspace{-0.7mm}
\includegraphics[width=.14\linewidth]{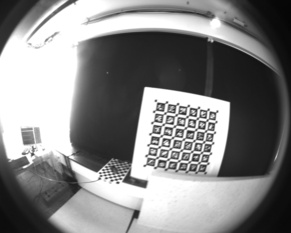}\hspace{-0.7mm}\\
\includegraphics[width=.14\linewidth]{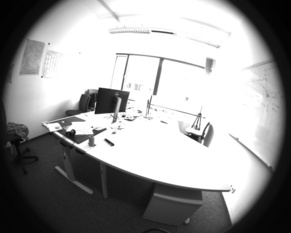}\hspace{-0.7mm}
\includegraphics[width=.14\linewidth]{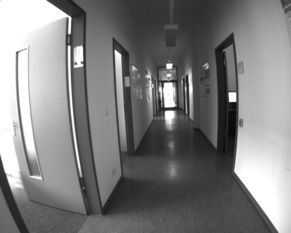}\hspace{-0.7mm}
\includegraphics[width=.14\linewidth]{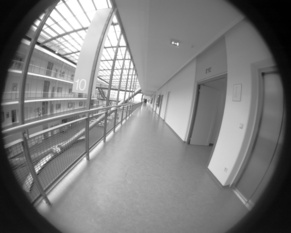}\hspace{-0.7mm}
\includegraphics[width=.14\linewidth]{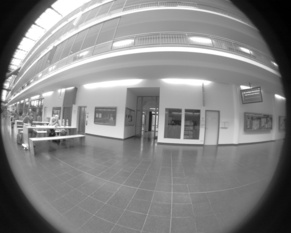}\hspace{-0.7mm}
\includegraphics[width=.14\linewidth]{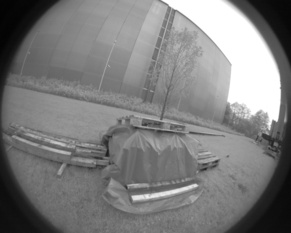}\hspace{-0.7mm}
\includegraphics[width=.14\linewidth]{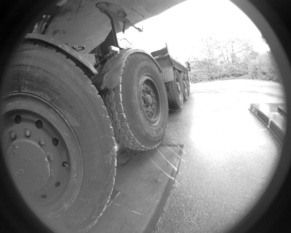}\hspace{-0.7mm}
\includegraphics[width=.14\linewidth]{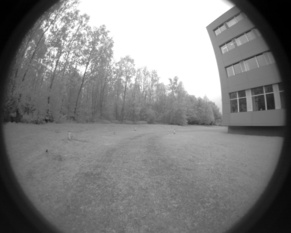}\hspace{-0.7mm}\\
\includegraphics[width=.14\linewidth]{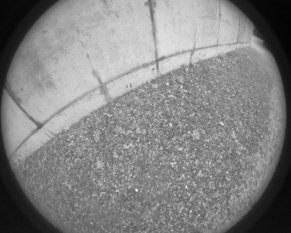}\hspace{-0.7mm}
\includegraphics[width=.14\linewidth]{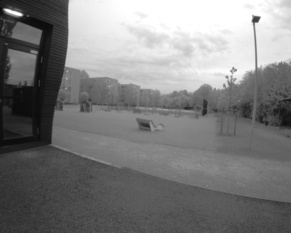}\hspace{-0.7mm}
\includegraphics[width=.14\linewidth]{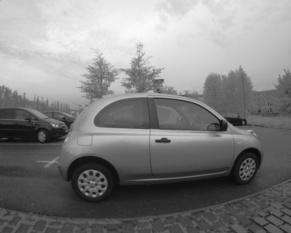}\hspace{-0.7mm}
\includegraphics[width=.14\linewidth]{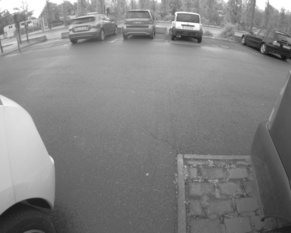}\hspace{-0.7mm}
\includegraphics[width=.14\linewidth]{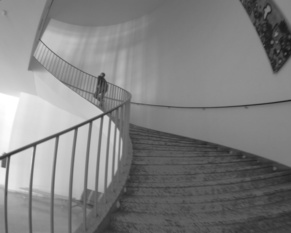}\hspace{-0.7mm}
\includegraphics[width=.14\linewidth]{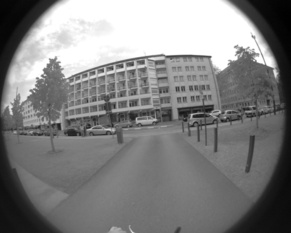}\hspace{-0.7mm}
\includegraphics[width=.14\linewidth]{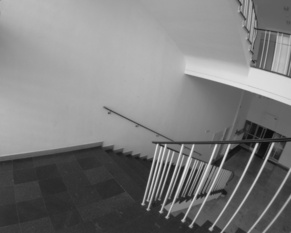}\hspace{-0.7mm}\\
\includegraphics[width=.14\linewidth]{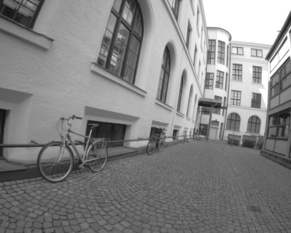}\hspace{-0.7mm}
\includegraphics[width=.14\linewidth]{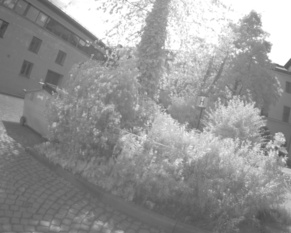}\hspace{-0.7mm}
\includegraphics[width=.14\linewidth]{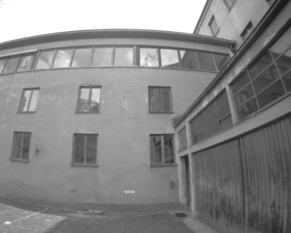}\hspace{-0.7mm}
\includegraphics[width=.14\linewidth]{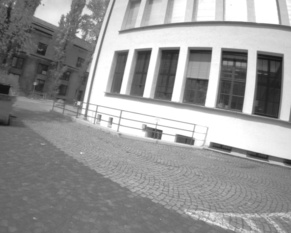}\hspace{-0.7mm}
\includegraphics[width=.14\linewidth]{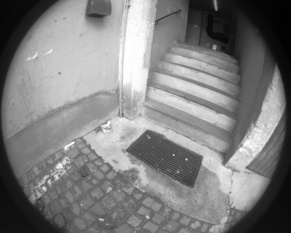}\hspace{-0.7mm}
\includegraphics[width=.14\linewidth]{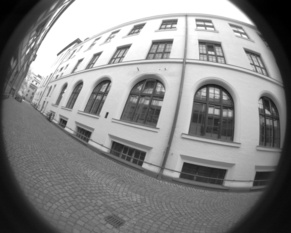}\hspace{-0.7mm}
\includegraphics[width=.14\linewidth]{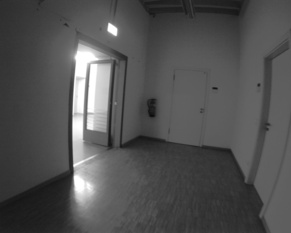}\hspace{-0.7mm}\\
\includegraphics[width=.14\linewidth]{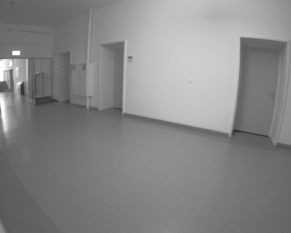}\hspace{-0.7mm}
\includegraphics[width=.14\linewidth]{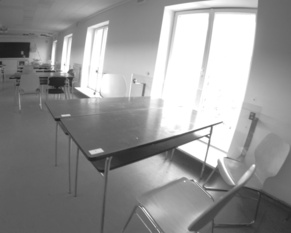}\hspace{-0.7mm}
\includegraphics[width=.14\linewidth]{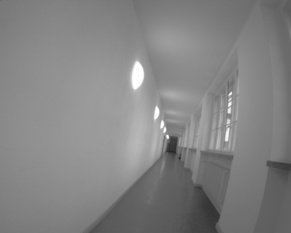}\hspace{-0.7mm}
\includegraphics[width=.14\linewidth]{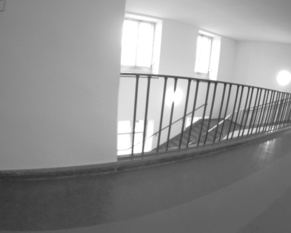}\hspace{-0.7mm}
\includegraphics[width=.14\linewidth]{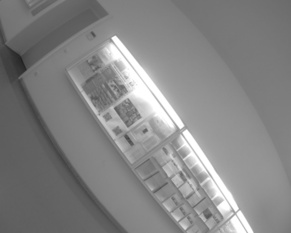}\hspace{-0.7mm}
\includegraphics[width=.14\linewidth]{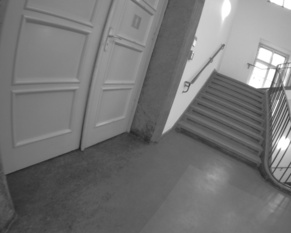}\hspace{-0.7mm}
\includegraphics[width=.14\linewidth]{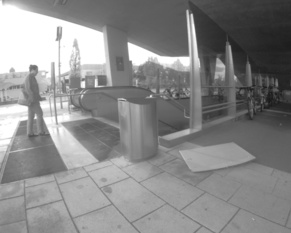}\hspace{-0.7mm}\\
\includegraphics[width=.14\linewidth]{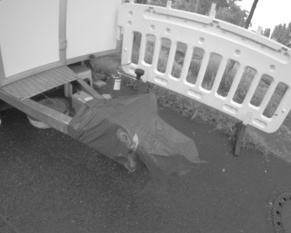}\hspace{-0.7mm}
\includegraphics[width=.14\linewidth]{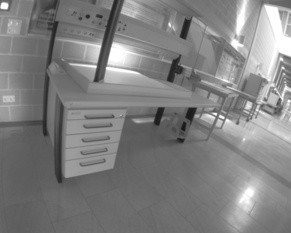}\hspace{-0.7mm}
\includegraphics[width=.14\linewidth]{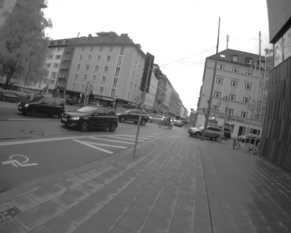}\hspace{-0.7mm}
\includegraphics[width=.14\linewidth]{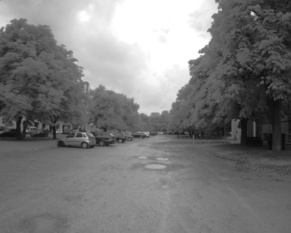}\hspace{-0.7mm}
\includegraphics[width=.14\linewidth]{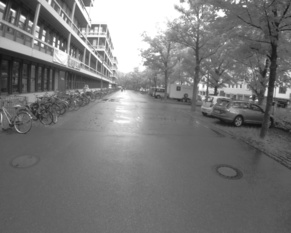}\hspace{-0.7mm}
\includegraphics[width=.14\linewidth]{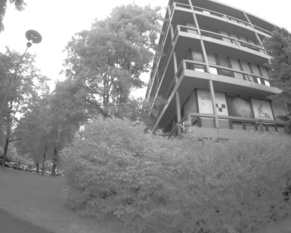}\hspace{-0.7mm}
\includegraphics[width=.14\linewidth]{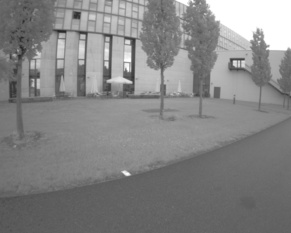}\hspace{-0.7mm}\\
\caption{\textbf{The TUM monoVO dataset.} A single frame from each of the 50 sequences.
Note the wide variety of covered environments.
The full dataset contains over 190'000 frames (105 minutes) of video taken with 
two different lenses, exposure times for each frame, and a photometric 
calibration including camera response and vignetting.}
\label{fig:teaser}
\end{figure}

\paragraph{Sensors Intrinsics.}
Many existing methods are designed to operate on, and are evaluated 
with, data captured by commodity cameras without taking advantage of knowing -- or even 
being able to influence -- the full image formation pipeline. 
Specifically, methods are designed to be robust to (assumed unknown) automatic exposure changes, 
non-linear response functions (gamma correction), lens attenuation (vignetting), de-bayering artifacts,
or even strong geometric distortions caused by a rolling shutter.
This is particularly true for modern keypoint detectors and descriptors, which are robust or invariant to 
arbitrary monotonic brightness changes. However, recent direct methods as well attempt to compensate 
for automatic exposure changes, e.g., by optimizing an affine mapping between brightness 
values in different images \cite{engel15iros}. Most direct methods however simply assume constant exposure time \cite{engel14eccv,newcombe2011iccv,pizzoli14icra,kerl13iros}.

While this is the only way to evaluate on existing datasets and with off-the-shelf commodity cameras 
(which often do not allow to either read or set parameters like the exposure time), 
we argue that for the above-mentioned use cases, this ultimately is the wrong approach: 
sensors -- including cameras -- can, and will be designed to fit the needs of the algorithms 
processing their data. In turn, algorithms should take full advantage of the 
sensor's capabilities and incorporate knowledge about the sensor design. Simple examples are image exposure time 
and hardware gamma correction, which are intentionally built into the camera to produce better images.
Instead of treating them as unknown noise factors and attempt to correct for them afterwards, 
they can be treated as feature that can be modelled by, and incorporated into the algorithm --
rendering the obtained data more meaningful.

\paragraph{Benchmark Size.}
SLAM and VO are complex, very non-linear estimation problems and often minuscule changes can greatly affect the outcome.
To obtain a meaningful comparison between different methods and to avoid manual overfitting 
to specific environments or motion patterns (except for cases where this is specifically desired), 
algorithms should be evaluated on large datasets in a wide variety of scenes.
However, existing datasets often contain only a limited number of 
environments. The major reason for this is that accurate ground truth acquisition is
challenging, in particular if a wide range of different environments is to be covered:
GPS/INS is limited in accuracy and only possible in outdoor environments with adequate GPS reception. External motion 
capture systems on the other hand are costly and time-consuming to set up, and can only cover small (indoor) environments. \\

The dataset published in this paper attempts to tackle these two issues. 
First, it contains frame-wise exposure times as reported by the sensor, as well as accurate calibrations for the sensors response 
function and lens vignetting, which enhances the performance particularly of direct approaches.
Second, it contains 50
sequences with a total duration of 105 minutes (see Figure~\ref{fig:teaser}), captured in dozens of different environments.
To make this possible, we propose a new evaluation methodology which does not require ground truth from 
external sensors -- instead, tracking accuracy is evaluated by measuring the accumulated drift that occurs
after a large loop.
We further propose a novel, straight-forward approach to calibrate a non-parametric response function and vignetting map with minimal
set up required, and without imposing a parametric model which may not suit all lenses / sensors.

\subsection{Related Work: Datasets}
There exists a number of datasets that can be used for evaluating monocular SLAM or
VO methods. We will here list the most commonly used ones.\\[-0.8cm]

\paragraph{KITTI \cite{geiger12cvpr}:} 21 stereo sequences recorded from a driving car, motion patterns and 
environments are limited to forward-motion and street-scenes. Images are pre-rectified,
raw sensor measurements or calibration datasets are not available. 
The benchmark contains GPS-INS ground truth poses for all frames.\\[-0.8cm]

\paragraph{EUROC MAV \cite{burrii6ijrr}:} 11 stereo-inertial sequences from a flying quadrocopter in three 
different indoor environments. The benchmark contains ground truth poses for all frames, as well as the raw 
sensor data and respective calibration datasets.\\[-0.8cm]

\paragraph{TUM RGB-D \cite{sturm12iros}:} 89 sequences in different categories (not all meant for SLAM) in various environments, recorded with a commodity RGB-D sensor.
They contain strong motion blur and rolling-shutter artifacts, as well as degenerate (rotation-only) motion patterns
that cannot be tracked well from monocular odometry alone. Sequences are pre-rectified, the raw sensor data is not available.
The benchmark contains ground truth poses for all sequences.\\[-0.8cm]

\paragraph{ICL-NUIM \cite{handa14icra}:} 8 ray-traced RGB-D sequences from 2 different environments. It provides
a ground truth intrinsic calibration; a photometric calibration is not required, as the virtual exposure time is constant.
Some of the sequences contain degenerate (rotation-only) motion patterns that cannot be tracked well from a monocular camera alone.

\subsection{Related Work: Photometric Calibration}
Many approaches exist to calibrate and remove vignetting artefacts and account for non-linear response functions. 
Early work focuses on image stitching and mosaicking, where the required calibration parameters
need to be estimated from a small set of overlapping images \cite{goldman05iccv} \cite{kim2008pami} \cite{brown2007ijcv}. 
Since the available data is limited, such methods attempt to find low-dimensional (parametric) function 
representations, like radially symmetric polynomial representations for vignetting.
More recent work \cite{alexandrov16iros,kerl143dv} has shown that such representations may not be sufficiently expressive to capture the complex 
nature of real-world lenses and hence advocate 
non-parametric -- dense -- vignetting calibration. In contrast to \cite{alexandrov16iros,kerl143dv}, our formulation however does not require a 
\quotes{uniformly lit white paper}, simplifying the required calibration set-up.

For response function estimation, a well-known and straight-forward method is that of Debevec and Malik \cite{debevec2008siggraph}, 
which -- like our approach -- recovers a $2^8$-valued lookup table for the inverse response from two or more images of a static
scene at different exposures.

\subsection{Paper Outline}
The paper is organized as follows:
In Section~\ref{sec:Calibration}, we first describe the hardware set-up, followed by 
both the used distortion model (geometric calibration) in \ref{ssec:CalibGeo}, as well as
photometric calibration (vignetting and response function) and the proposed calibration procedure in \ref{ssPhotoCalib}. 
Section~\ref{sec:EvaluationMetric} describes the proposed loop-closure evaluation methodology and respective error measures.
Finally, in Section~\ref{sec:Benchmark}, we give extensive evaluation results of two state-of-the-art monocular SLAM / VO systems
ORB-SLAM \cite{mur2015orb} and Direct Sparse odometry (DSO) \cite{engel16archiveOdometry}. We further show some exemplary image data and describe the dataset contents.
In addition to the dataset, we publish all code and raw evaluation data as open-source.

\begin{figure}
\centering
\includegraphics[width=.99\linewidth]{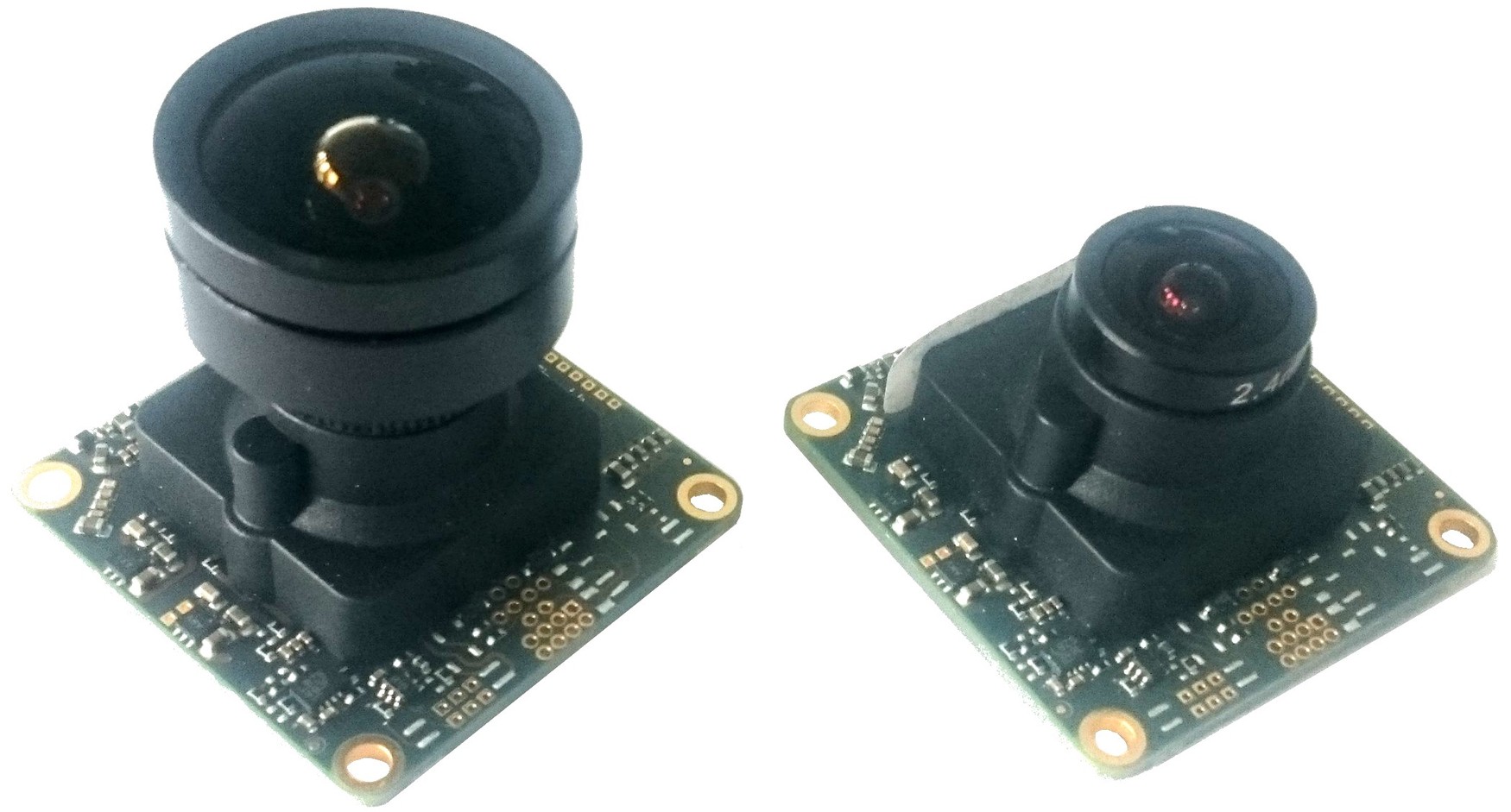}\hspace{-0.7mm}
\caption{\textbf{Cameras used to capture the dataset.} Left: narrow lens ($98^\circ \times 79^\circ$ non-rectified field of view), 
right: wide lens ($148^\circ \times 122^\circ$ non-rectified field of view).}
\label{fig:cameras}
\end{figure}

\section{Calibration}
\label{sec:Calibration}
We provide both a standard camera intrinsic calibration using the FOV camera model,
as well as a photometric calibration, including vignetting and camera response function.

\subsection{Hardware}
\label{ssec:Hardware}
The cameras used for recording the sequences are uEye UI-3241LE-M-GL monochrome, global shutter CMOS cameras from IDS. 
They are capable of recording $1280 \times 1024$ videos at up to 60fps. Sequences are recorded at different framerates ranging
from 20fps to 50fps with jpeg-compression. For some sequences hardware gamma correction is enabled and for some it is disabled. 
We use two different lenses (Lensagon BM2420 with a field of view of $148^\circ \times 122^\circ$, as well as a 
Lensagon BM4018S118 with a field of view of $98^\circ \times 79^\circ$), as shown in 
Figure~\ref{fig:cameras}. Figure~\ref{fig:teaser} shows a number of example images from the dataset.

\subsection{Geometric Intrinsic Calibration}
\label{ssec:CalibGeo}
We use the pinhole camera model in combination with a FOV distortion model, since 
it is well-suited for the used fisheye lenses. For a given 3D point $(x,y,z) \in \mathbb{R}^3$
in the camera coordinate system, the corresponding point in the image $(u_d, v_d) \in \Omega$ is computed by first applying
a pinhole projection
followed by radial distortion and conversion to pixel coordinates
\begin{align}
	\begin{bmatrix}u_d\\v_d\end{bmatrix} = \frac{1}{r_u \omega } \arctan \left( 2 r_u \tan\left(\frac{\omega}{2}\right)\right) \begin{bmatrix}f_x \frac{x}{z}\\f_y \frac{y}{z}\end{bmatrix}
	+ \begin{bmatrix}c_x\\c_y\end{bmatrix},
\end{align}
where $r_u := \sqrt{(\frac{x}{z})^2 + (\frac{y}{z})^2}$ is the radius of the point in normalized image coordinates.

A useful property of this model is the existence of a closed-form inverse: for a given point 
the image $(u_d, v_d)$ and depth $d$, the corresponding 3D point can be computed by first converting it back to normalized image coordinates
\begin{align}
	\begin{bmatrix}\tilde{u}_d\\\tilde{v}_d\end{bmatrix} = \begin{bmatrix}(u_d-c_x) f_x^{-1}\\(v_d-c_y) f_y^{-1}\end{bmatrix},
\end{align}
then removing radial distortion
\begin{align}
	\begin{bmatrix}\tilde{u}_u\\\tilde{v}_u\end{bmatrix} = \frac{\tan(r_d \omega)}{2 r_d \tan \frac{\omega}{2}} \begin{bmatrix}\tilde{u}_d\\\tilde{v}_d\end{bmatrix}, 
\end{align}
where $r_d := \sqrt{\tilde{u}_d^2 + \tilde{v}_d^2}$.
Afterwards the point is back-projected using 
$(x,y,z) = d (\tilde{u}_u,\tilde{v}_u,1)$.
We use the camera calibrator provided with the open-source implementation of PTAM \cite{klein07ismar} to calibrate the 
parameters $[f_x, f_y, c_x, c_y, \omega]$ from a number of checkerboard images.

\subsection{Photometric Calibration}
\label{ssPhotoCalib}
We provide photometric calibrations for all sequences. 
We calibrate the camera response function $G$, as 
well as pixel-wise attenuation factors $V \colon \Omega \to [0,1]$ (vignetting).
Without known irradiance, both $G$ and $V$ are only observable up to a scalar factor.
The combined image formation model is then given by
\begin{align}
	I(\mathbf{x}) = G\big(t V(\mathbf{x}) B(\mathbf{x})\big),
\end{align}
where $t$ is the exposure time, $B$ the irradiance image (up to a scalar factor), and $I$ the 
observed pixel value. As a shorthand, we will use $U:=G^{-1}$ for the inverse response function.

\begin{figure}
\centering
\includegraphics[width=.19\linewidth]{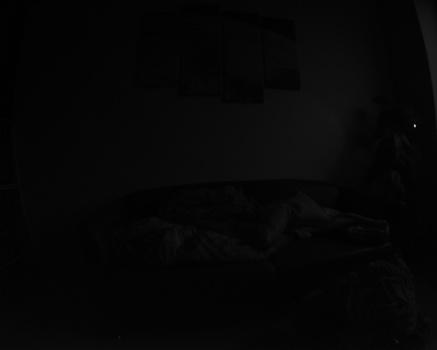}
\includegraphics[width=.19\linewidth]{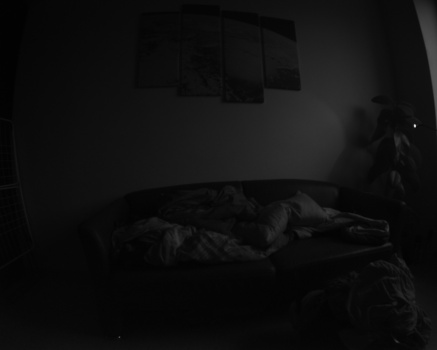}
\includegraphics[width=.19\linewidth]{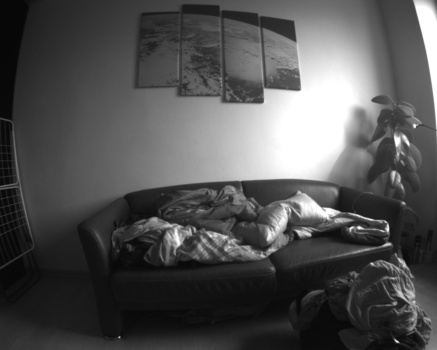}
\includegraphics[width=.19\linewidth]{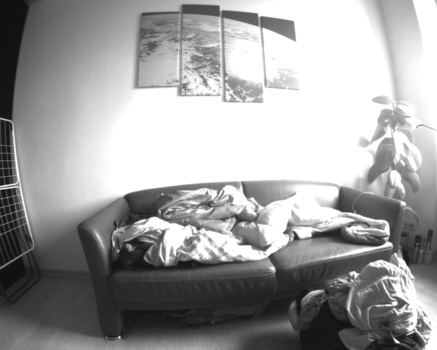}
\includegraphics[width=.19\linewidth]{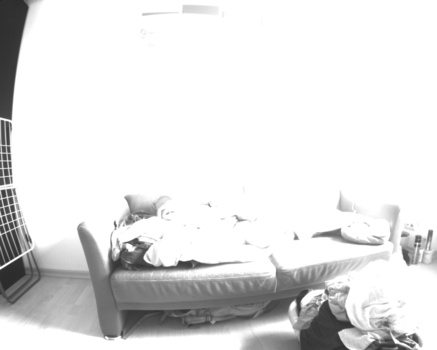}\\[1mm]
\begin{minipage}{0.50\linewidth}\includegraphics[width=1\linewidth]{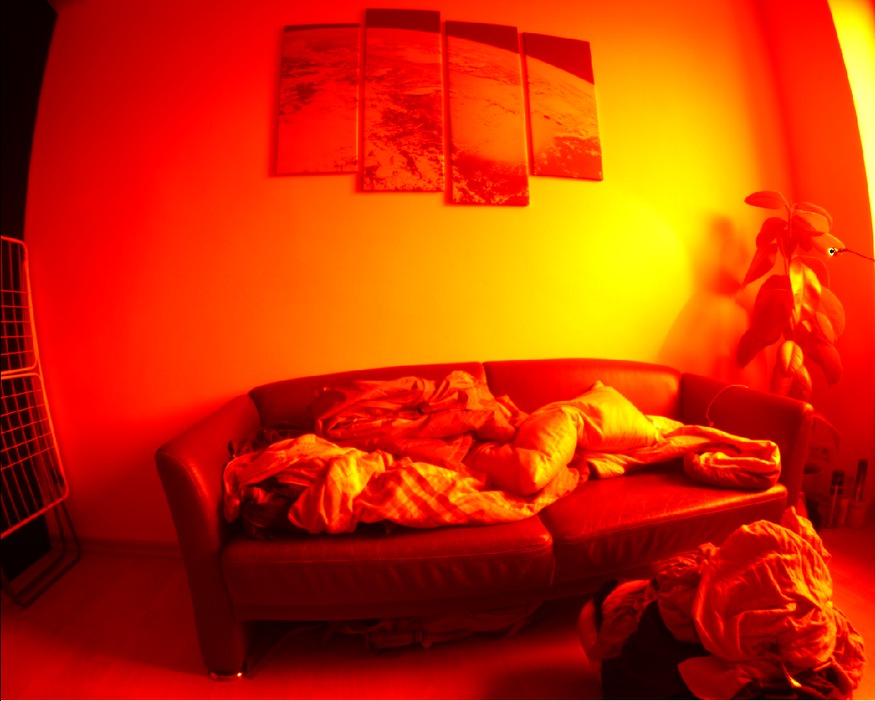}\end{minipage}
\hspace{1.5mm}
\begin{minipage}{0.03\linewidth}\rotatebox{90}{\footnotesize ~~~irradiance $U(I)$}\\\end{minipage}
\begin{minipage}{0.41\linewidth}\includegraphics[width=1\linewidth]{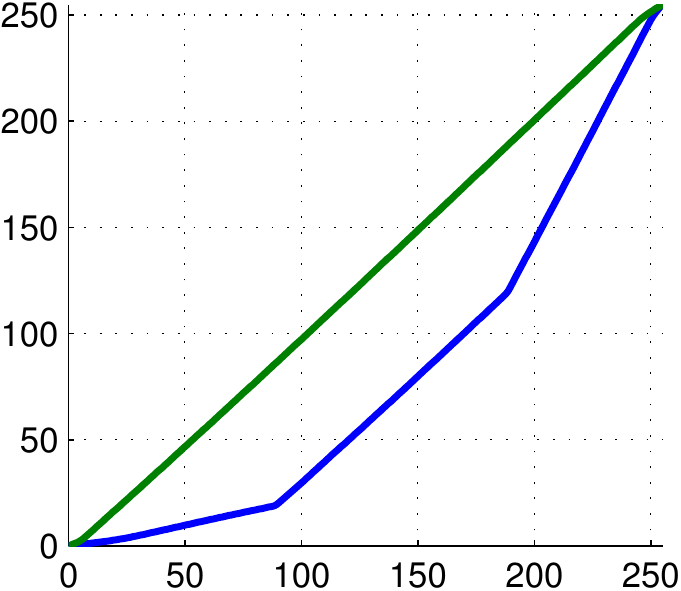}\\[-1mm]\footnotesize \hspace*{1.3cm}pixel value $I$\end{minipage}\\[1mm]
\caption{\textbf{Response calibration.} The top row shows five out of over 1000 images of the
same scene at different exposures, used for calibration.
The bottom-left image shows the estimated log-irradiance $\log (B')$; the bottom-right image shows the estimated 
inverse response $U$, with enabled hardware gamma correction (blue) and without (green).}
\label{fig:responseCalib}
\end{figure}

\subsubsection{Response Calibration}
\label{sssec:ResponseCalib}
We first calibrate the camera response function from a sequence of images taken of a
static scene with different (known) exposure time. The content of the scene is arbitrary --
however to well-constrain the problem, it should contain a wide range of gray values.
We first observe that in a static scene, the attenuation 
factors can be absorbed in the irradiance image, i.e.,
\begin{align}
	I(\mathbf{x}) = G\big(t B'(\mathbf{x})\big),
\end{align}
with \mbox{$B'(\mathbf{x}):= V(\mathbf{x}) B(\mathbf{x})$}. Given a number of images 
$I_i$, corresponding exposure times $t_i$ and a Gaussian white noise assumption 
on $U(I_i(\mathbf{x}))$, this leads to the following Maximum-Likelihood energy formulation
\begin{align}
\label{eqGamma}
	E(U, B') = \sum_i \sum_{\mathbf{x} \in \Omega} \Big(U\big(I_i(\mathbf{x})\big) - t_i B'(\mathbf{x})\Big)^2.
\end{align}
For overexposed pixels, $U$ is not well defined, hence they are removed from the estimation.
We now minimize (\ref{eqGamma}) alternatingly for $U$ and $B'$. 
Note that fixing either $U$ or $B'$ de-couples all remaining values, such that 
minimization becomes trivial:
\begin{align}
	U(k)^* & = \argmin_{U(k)} E(U, B') 
		= \frac{\sum_{\Omega_{k}} t_i B'(\mathbf{x})}{|\Omega_{k}|}\\
	B'(\mathbf{x})^* & = \argmin_{B'(\mathbf{x})} E(U, B')
	 	= \frac{\sum_i t_i U\big(I_i(\mathbf{x})\big)}{\sum_i t_i^2},
\end{align}
where $\Omega_{k} := \{ i, \mathbf{x} | I_i(\mathbf{x})=k \}$ is the set of all pixels in all images that have intensity $k$. 
Note that the resulting $U$ may not be monotonic -- which is a pre-requisite for invertibility. In 
this case it needs to be smoothed or perturbed; however for all our calibration datasets this does not happen.
The value for $U(255)$ is never observed since overexposed pixels are removed, and needs to be extrapolated from the adjacent values.
After minimization, $U$ is scaled such that $U(255)=255$ to disambiguate the unknown scalar factor.
Figure \ref{fig:responseCalib} shows the estimated values for one of the calibration sequences.

\paragraph{Note on Observability.} In contrast to \cite{debevec2008siggraph}, we do not employ a smoothness prior on $U$ -- instead, we 
use large amounts of data (1000 images covering 120 different exposure times, ranging from 0.05ms to 20ms in 
multiplicative increments of 1.05). This is done by recording a video of a static scene 
while slowly changing the camera's exposure. If only a small number of images or
exposure times is available, a regularized approach will be required.

\begin{figure}
\centering
\begin{minipage}{.55\linewidth}
\includegraphics[width=.49\linewidth]{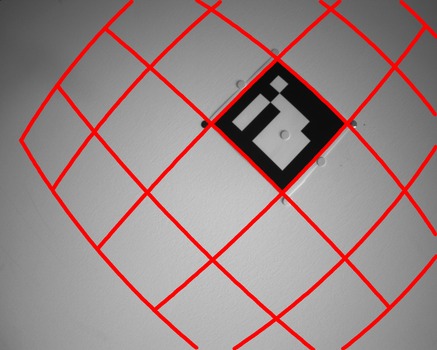}\hspace{-0.4mm}
\includegraphics[width=.49\linewidth]{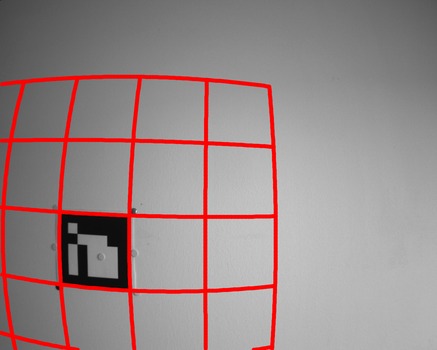}\\
\includegraphics[width=.49\linewidth]{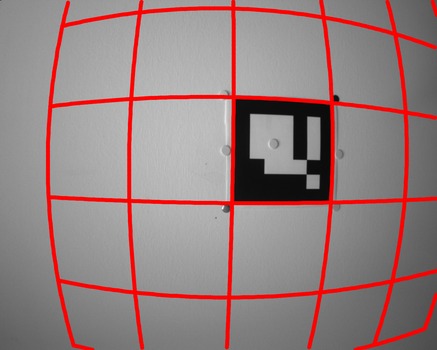}\hspace{-0.4mm}
\includegraphics[width=.49\linewidth]{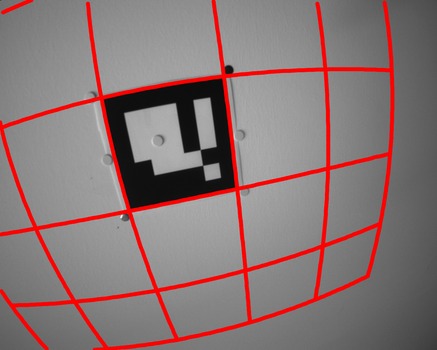}
\end{minipage}
\begin{minipage}{.44\linewidth}
\includegraphics[width=.99\linewidth]{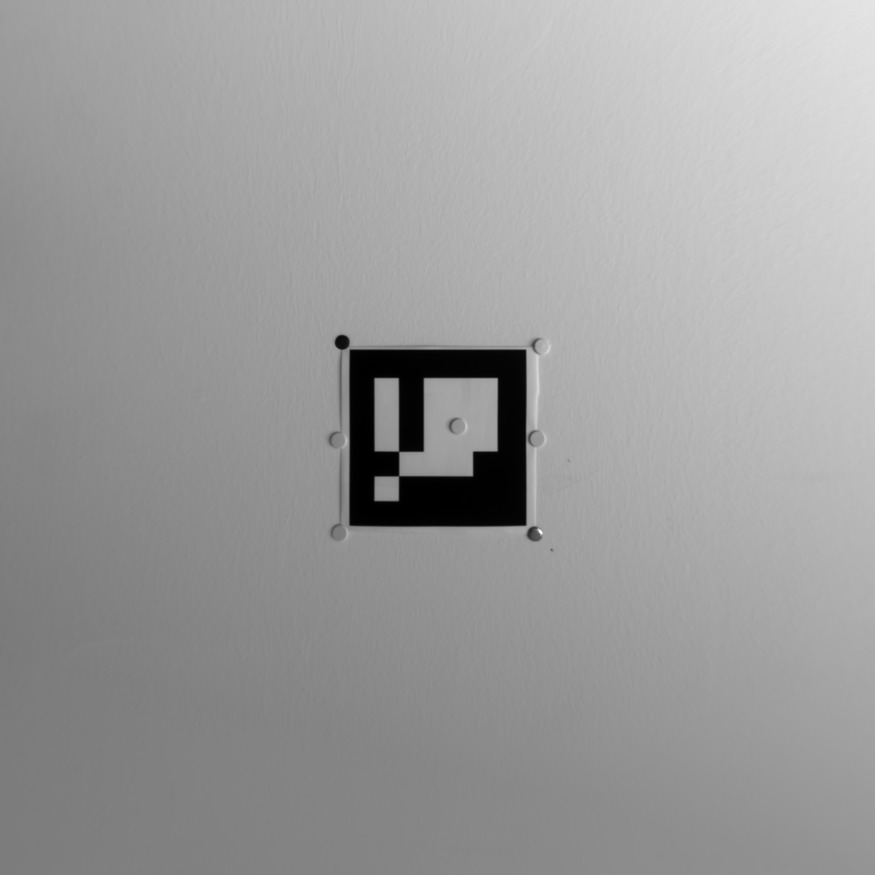}
\end{minipage}\\[.5mm]
\includegraphics[width=\linewidth]{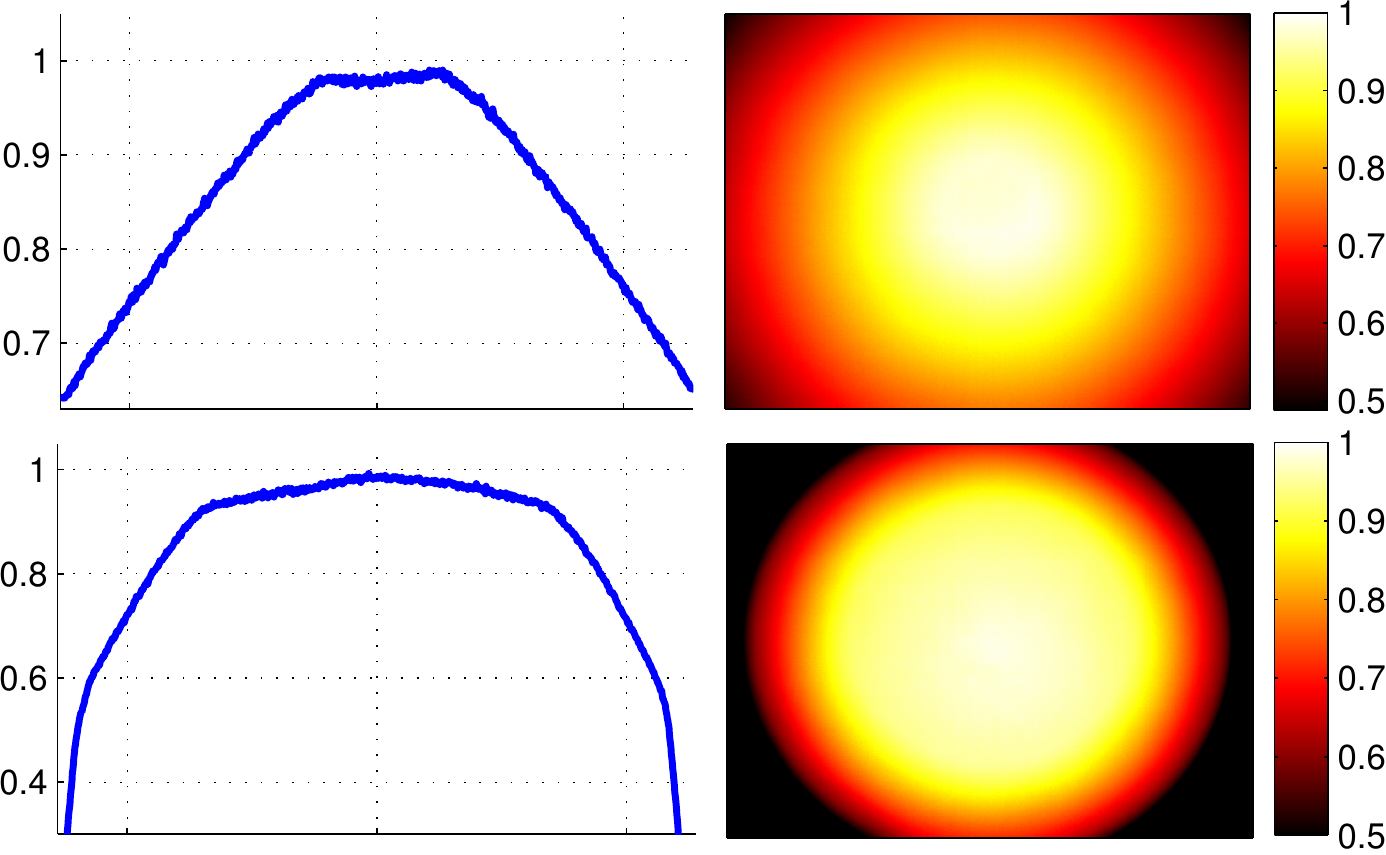}
\caption{\textbf{Vignette calibration.} Top-left: four out of over 700 images used for vignette calibration,
overlaid with the 3D plane $\mathcal{P}$ in red. Top-right: estimated irradiance image $C$ for plane $\mathcal{P}$. 
Bottom-right: estimated dense attenuation factors $V$, for both used lenses.  Bottom-left: horizontal cross-section through
$V$ at the middle of the image.}
\label{fig:vignette}
\end{figure}

\subsubsection{Non-parametric Vignette Calibration}
\label{sssec:VignetteCalib}
We estimate a non-parametric (dense) vignetting map $V \colon \Omega \to [0,1]$ from a sequence 
of images showing a planar scene. Apart from planarity, we only require the scene to have a bright 
(potentially non-uniform) color and to be fully Lambertian -- in practice, a predominantly white wall serves well. 
This is in contrast to \cite{alexandrov16iros}, which assumes 
a uniformly coloured flat surface. For simplicity, we estimate the camera pose with 
respect to the planar surface $\mathcal{P} \subset \mathbb{R}^3$ using an AR Marker \cite{aruco2014}; 
however any other method, including a full monocular SLAM system can be used.
For each image $I_i$, this results in a mapping $\pi \colon \mathcal{P} \to \Omega$ that projects a 
point on the 3D plane to a pixel in the image (if it is visible).

Again, we assume Gaussian white noise on $U(I_i(\pi_i(\mathbf{x})))$, leading to the Maximum-Likelihood energy
\begin{align}
\label{eqEnergyVignette}
	E(C, V)\!=\!\sum_{i, \mathbf{x} \in  \mathcal{P}}\!\bigg( t_i V\Big(\big[\pi_i(\mathbf{x})\big]\Big) C(\mathbf{x}) - U\Big(I_i\big(\pi_i(\mathbf{x})\big)\Big) \bigg)^2,
\end{align}
where $C \colon \mathcal{P} \to \mathbb{R}$ is the unknown irradiance of the planar surface. In practice we define the 
surface to be square, and discretise it into $1000 \times 1000$ points; $[\cdot]$ then denotes 
rounding to the closest discretised position. 

The energy $E(C, V)$ can be minimized alternatingly, again fixing one variable decouples all other 
unknowns, such that minimization becomes trivial:
\begin{align}
	C^*(\mathbf{x}) &= \argmin_{C(\mathbf{x})} E(C,V) \nonumber \\
	& = \frac{\sum_i t_i V\big([\pi_i(\mathbf{x})]\big) U\big(I_i(\pi_i(\mathbf{x}))\big)}{\sum_i \big( t_i V([\pi_i(\mathbf{x})]) \big)^2},\\
	V^*(\mathbf{x}) &= \argmin_{V(\mathbf{x})} E(C,V) \nonumber \\
	& = \frac{\sum_i t_i C(\mathbf{x}) U\big(I_i(\pi_i(\mathbf{x}))\big)}{\sum_i \big( t_i C(\mathbf{x}) \big)^2}.
\end{align}
Again, we do not impose any explicit smoothness prior or enforce certain 
properties (like radial symmetry) by finding a parametric representation for $V$. Instead, we
choose to solve the problem by using large amounts (several hundred images) of data. Since $V$ is
only observable up to a scalar factor, we scale the result such that $\max(V) = 1$.
Figure \ref{fig:vignette} shows the estimated attenuation factor map for both lenses. The high degree of radial 
symmetry and smoothness comes solely from the data term, without additional prior.

\paragraph{Note on Observability.} Without regularizer, the optimization problem (\ref{eqEnergyVignette}) is well-constrained if and only if the 
corresponding bipartite graph between all optimization variables is fully connected. 
In practice, the probability that this is not the case is negligible when using sufficient input images: 
Let $(A, B, E)$ be a random bipartite graph with $|A| = |B| = n$ nodes and $|E|=[n (\log n + c)]$ edges, 
where $c$ is a positive real number. We argue that the complex nature of 3D projection and perspective warping 
justifies the approximation of the resulting residual graph as random, provided the input images cover a wide range of viewpoints.
Using the Erd\H{o}s-R\'{e}nyi theorem \cite{erdos59graph}, it can be shown that 
for $n \to \infty$, the probability of the graph being connected is given by $P = e^{-2 e^{-c}}$ \cite{palasti1963}.
In our case, $n \approx 1000^2$, which is sufficiently large for this approximation to be good.
This implies that $30n$ residuals (i.e., 30 images with the full plane visible) suffice for the problem 
to be almost certainly well-defined ($P > 0.999999$).
To obtain a good solution and fast convergence, a significantly larger number of input images is desirable 
(we use several hundred images) which are easily captured by taking a short (60s) video, that
covers different image regions with different plane regions.

\section{Evaluation Metrics}
\label{sec:EvaluationMetric}

\begin{figure}
\centering
{\setlength{\fboxsep}{0pt}\setlength{\fboxrule}{0.4pt}
\fbox{\includegraphics[width=.49\linewidth]{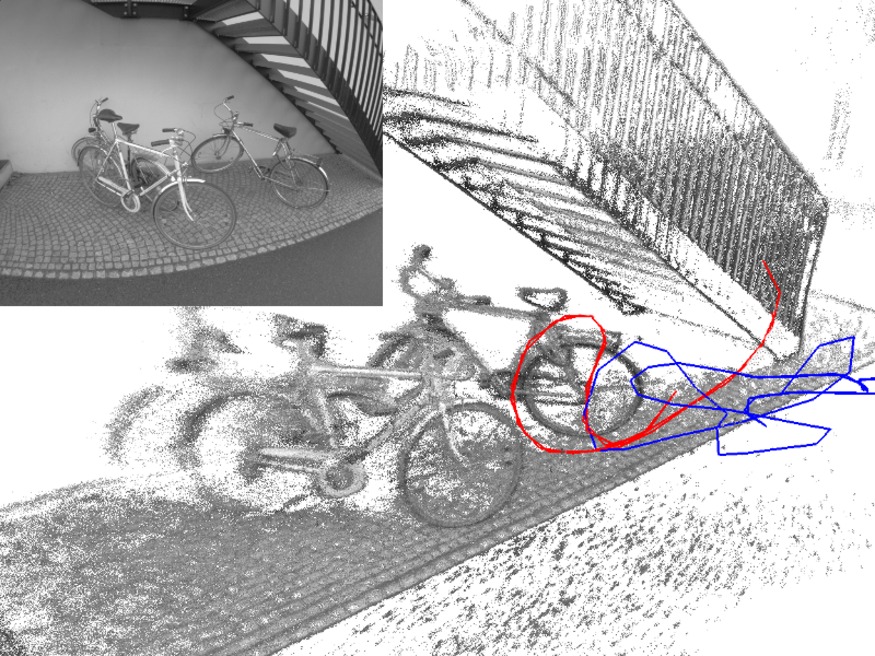}}\hspace{-0.4mm}
\fbox{\includegraphics[width=.49\linewidth]{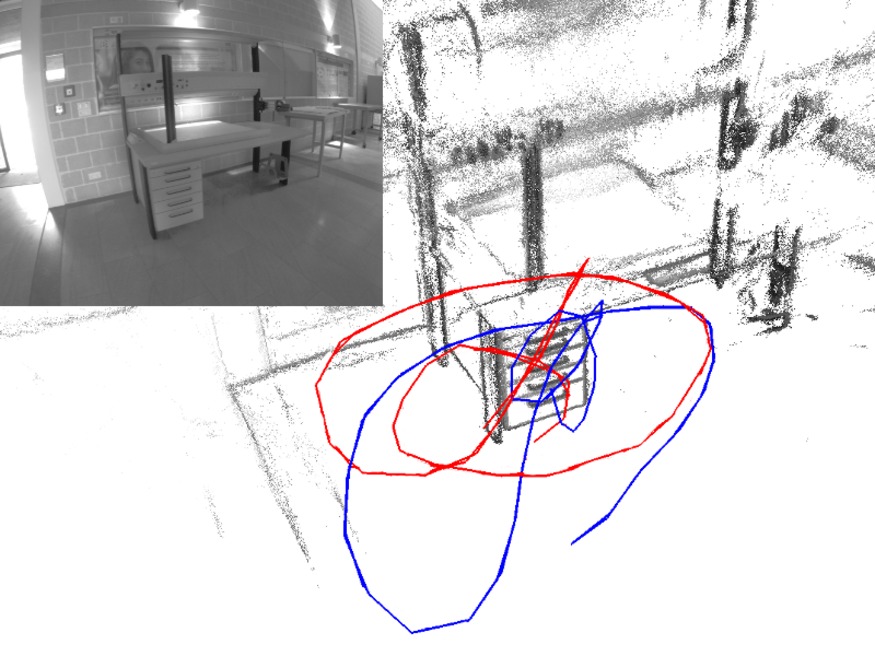}}\\
\fbox{\includegraphics[width=.49\linewidth]{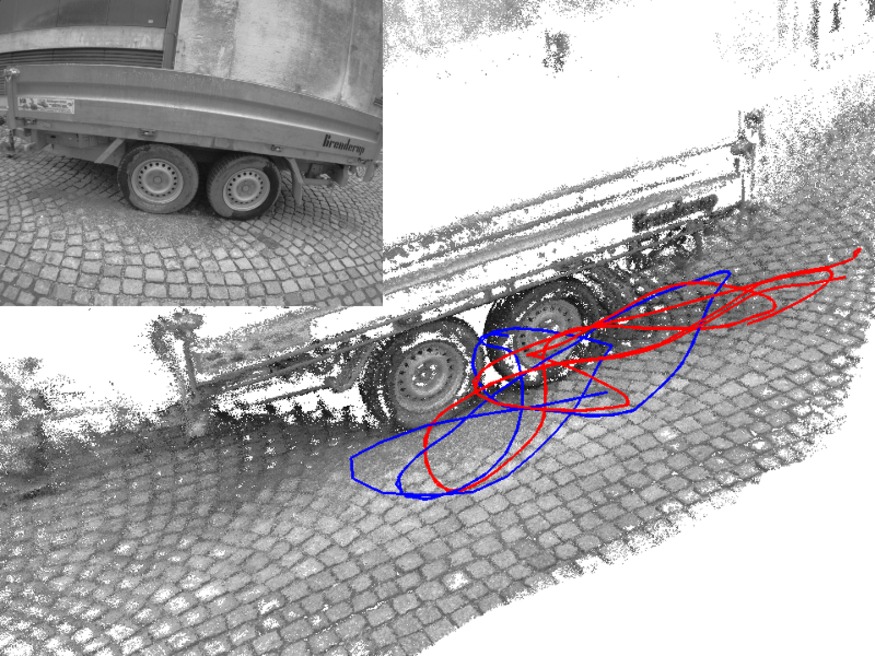}}\hspace{-0.4mm}
\fbox{\includegraphics[width=.49\linewidth]{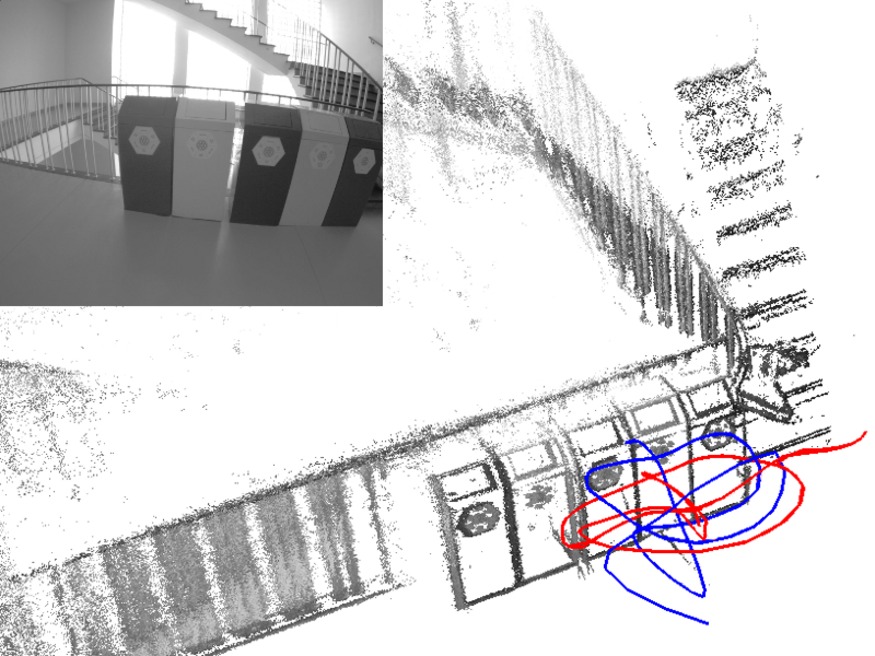}}}
\caption{\textbf{Loop-closure alignment.} Explicit loop-closure alignment for 4 
selected sequences, created with LSD-SLAM. The red and blue line correspond to 
the first and last segment of the full trajectory.}
\label{fig:loopclosures}
\end{figure}

\subsection{Evaluation from Loop-Closure}
The dataset focuses on a large variety of real-world indoor and outdoor scenes, for which it is
very difficult to obtain metric ground truth poses.
Instead, all sequences contain exploring motion, and have one large loop-closure at the end: 
The first and the last 10-20 seconds of each sequence show the same, easy-to-track scene, with slow, loopy camera motion. 
We use LSD-SLAM \cite{engel14eccv} to track only these segments, and -- very precisely -- align the 
start and end segment, generating a \quotes{ground truth} for their relative pose\footnote{Some 
sequences begin and end in a room which is equipped with a motion capture system. For those sequences, we use 
the metric ground truth from the MoCap.}. To provide better comparability between the sequences,
the ground truth scale is normalized such that the full trajectory has a length of approximately 100.

The tracking accuracy of a VO method can then be evaluated in terms of the accumulated error 
(drift) over the full sequence. \textbf{Note that this evaluation method is only valid if the VO/SLAM method does not
perform loop-closure itself. To evaluate full SLAM systems like ORB-SLAM or LSD-SLAM, loop-closure detection
needs to be disabled}. We argue that even for full SLAM methods, the amount of drift accumulated
before closing the loop is a good indicator for the accuracy. In particular, it is strongly correlated with the 
long-term accuracy after loop-closure.

It is important to mention that apart from accuracy, full SLAM includes a number of additional important challenges such as
loop-closure detection and subsequent map correction, re-localization, and long-term map maintenance 
(life-long mapping) -- all of which are not evaluated with the proposed set-up.

\begin{figure}
\centering
\includegraphics[width=.99\linewidth]{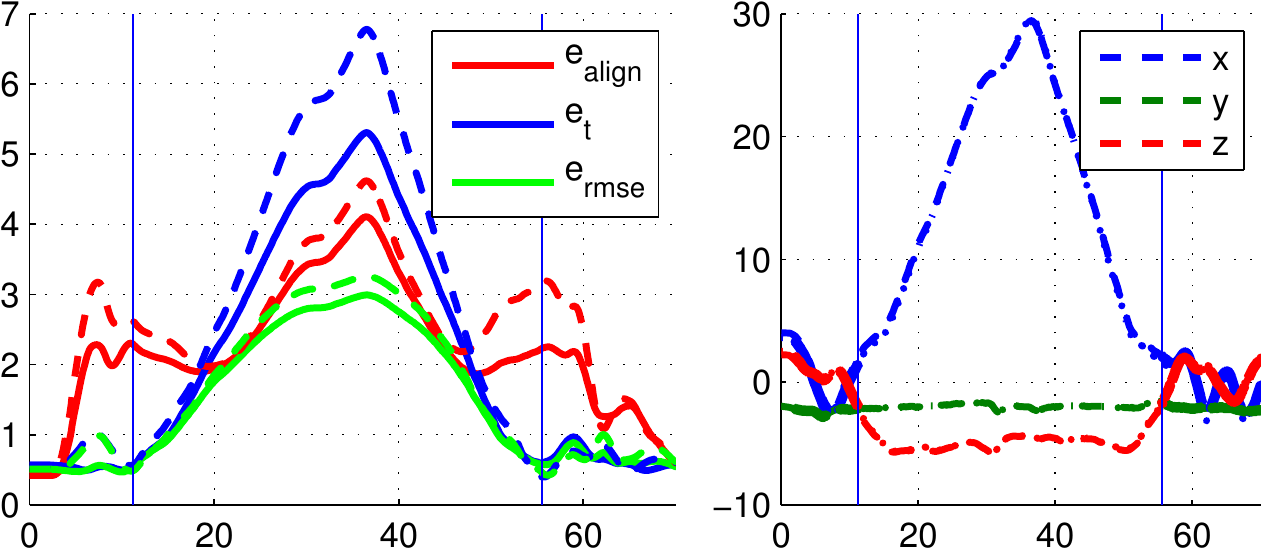}
\caption{\textbf{Evaluation metric.} The right plot shows a tracked $(x,y,z)$-trajectory for \texttt{sequence\_16}. We analyse the effect of drift on the different error
metrics by adding an artificial scale-jump of $\times 0.8$ or rotation-jump of $10^\circ$ at different time-points throughout the sequence: 
The left plot shows how the error metrics depend on the time-point where drift occurs (scale: dashed, rotation: solid). Both $e_t$ and $e_\text{rmse}$ are heavily correlated 
with \textit{where} the drift occurs (the further away, the more impact it has). The alignment error $e_\text{align}$ in contrast behaves much 
more stable, i.e., the error is less susceptible to where the drift occurs.}
\label{fig:metricCompare}
\end{figure}

\subsection{Error Metric}
Evaluation proceeds as follows: 
Let $p_1 \hdots p_n \in \mathbb{R}^3$ denote the tracked positions of frames
$1$ to $n$. Let  $S \subset [1; n]$ and $E \subset [1; n]$ be the 
frame-indices for the start- and end-segments for which aligned ground truth positions $\hat{p} \in \mathbb{R}^3$ are provided.
First, we align the tracked trajectory with both the start- and end-segment independently, providing two relative transformations
\begin{align}
\label{eq:AlignS}
	T_s^\text{gt} := \argmin_{T\in \text{Sim}(3)} \sum_{i \in S} (T p_i - \hat{p}_i)^2\\
\label{eq:AlignE}
	T_e^\text{gt} := \argmin_{T\in \text{Sim}(3)} \sum_{i \in E} (T p_i - \hat{p}_i)^2.
\end{align}
For this step it is important, that both $E$ and $S$ contain sufficient poses in a non-degenerate configuration to well-constrain the alignment -- hence the loopy motion patterns at the beginning and end of each sequence.
The accumulated drift can now be computed as $T_\text{drift} = T_e^\text{gt}(T_s^\text{gt})^{-1} \in \text{Sim}(3)$, from which we can explicitly compute
(a) the scale-drift $e_s := \text{scale}(T_\text{drift})$, (b) the rotation-drift $e_r := \text{rotation}(T_\text{drift})$ and (c) the translation-drift $e_t := \|\text{translation}(T_\text{drift})\|$.

\begin{figure}
\centering
{\setlength{\fboxsep}{0pt}\setlength{\fboxrule}{0.4pt}
\fbox{\includegraphics[width=.99\linewidth]{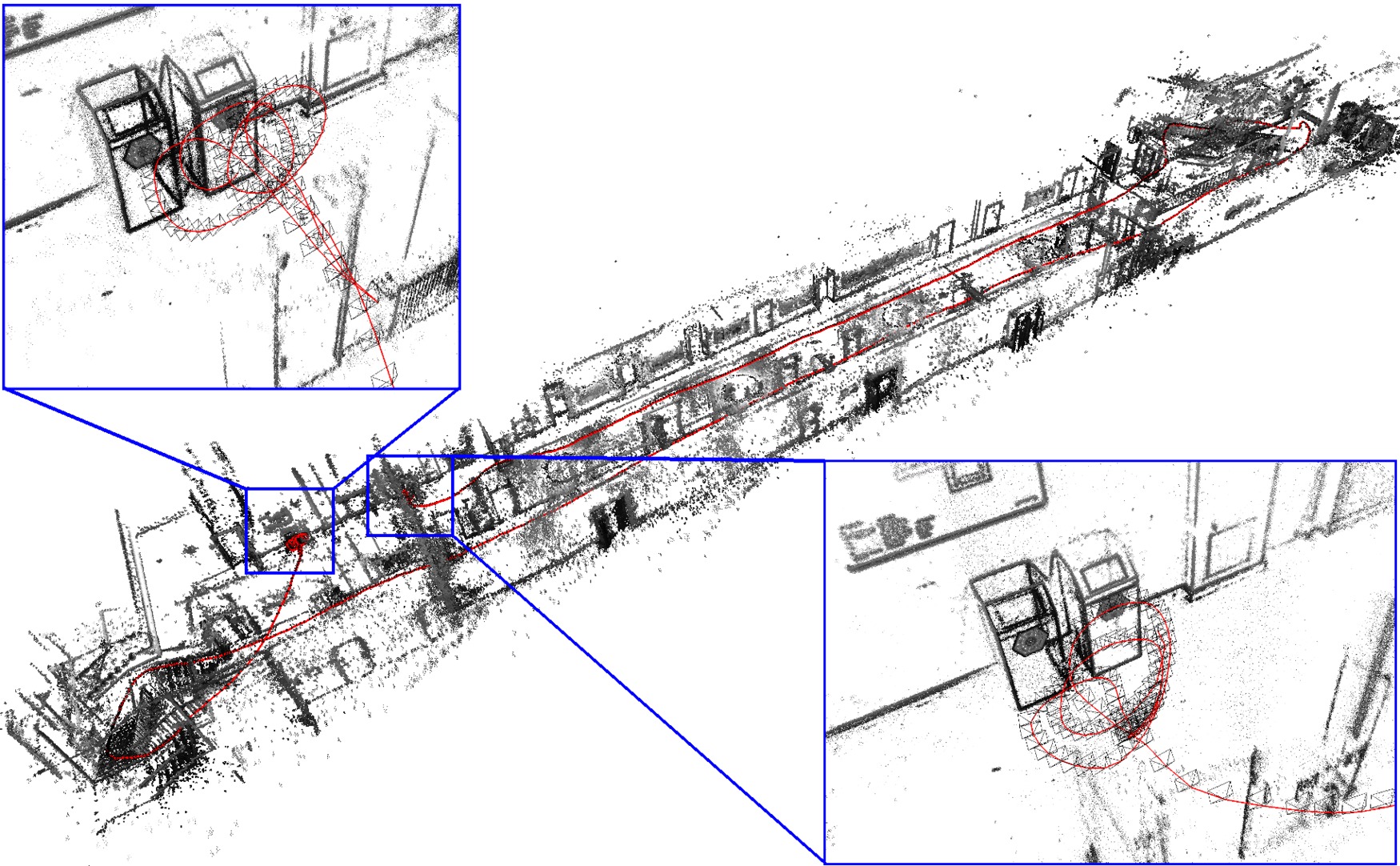}}\\
{\includegraphics[width=.99\linewidth]{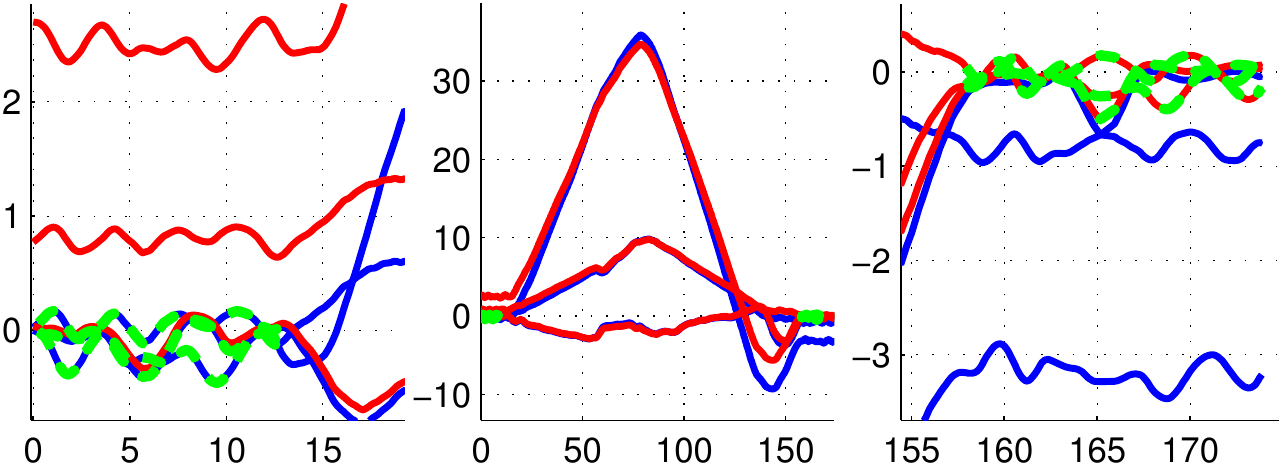}}}
\caption{\textbf{Alignment error.} The top shows \texttt{sequence\_40}, tracked using \cite{engel16archiveOdometry}. The accumulated drift
is clearly visible in the reconstruction (the two enlarged segments should overlap). The bottom plots show the loop-closure ground truth (dashed, green), 
and the tracked trajectory (1) aligned to the start-segment in blue and (2) aligned to the end segment in red (the center plot shows the full trajectory, the left and right plot show a close-up of the start- and end-segment respectively). The alignment error $e_\text{align}$ computes the RMSE between the red and the blue line, over the full trajectory. For this example, $e_\text{align}=2.27$, $e_\text{s} = 1.12$ and $e_\text{r}=3.9^\circ$.}
\label{fig:metricExample}
\end{figure}

We further define a combined error measure, the \textit{alignment error}, which equally takes into account the 
error caused by scale, rotation and translation drift over the full trajectory as
\begin{align}
	e_\text{align} := \sqrt{\frac{1}{n} \sum_{i=1}^n \|T_s^\text{gt} p_i - T_e^\text{gt} p_i\|_2^2},
\end{align}
which is the translational RMSE between the tracked trajectory, when aligned (a) to the 
start segment and (b) to the end segment. Figure \ref{fig:metricExample} shows an example. 
We choose this metric, since 
\begin{itemize}
	\item it can be applied to other SLAM / VO modes with different observability modes (like visual-inertial or stereo),
	\item it is equally affected by scale, rotation, and translation drift, implicitly weighted by their effect on the tracked position,
	\item it can be applied for algorithms which compute only poses for a subset of frames (e.g. keyframes), as long as start- and end-segment contain sufficient frames for alignment, and
	\item it better reflects the overall accuracy of the algorithm than the translational drift
drift $e_t$ or the joint RMSE
\begin{align}
\label{eqE_RMSE}
	e_\text{rmse}\!\!:=\!\!\sqrt{\min_{T\in \text{Sim}(3)} \frac{1}{|S \cup E|} \sum_{i \in S \cup E} (T p_i - \hat{p}_i)^2},
\end{align}
as shown in Figure~\ref{fig:metricCompare}. In particular, $e_\text{rmse}$ becomes degenerate for sequences where the 
accumulated translational drift surpasses the standard deviation of $\hat{p}$, since the alignment (\ref{eqE_RMSE}) will simply
optimize to \mbox{$\text{scale}(T) \approx 0$}.
\end{itemize}

\section{Benchmark}
\label{sec:Benchmark}
When evaluating accuracy of SLAM or VO methods, a common issue is that 
not all methods work on all sequences. This is particularely
true for monocular methods, as sequences with degenerate (rotation-only) motion
or entirely texture-less scenes (white walls) cannot be tracked.
All methods will then either produce arbitrarily bad (random) estimates
or heuristically decide they are \quotes{lost}, and not provide an estimate at all.
In both cases, averaging over a set of results containing such outliers is 
not meaningful, since the average will mostly reflect the (arbitrarily bad) errors
when tracking fails, or the threshold when the algorithm decides to not 
provide an estimate at all.

\begin{figure}
\centering
\begin{minipage}{0.03\linewidth}\rotatebox{90}{\footnotesize ~~~number of runs}\\\end{minipage}
\begin{minipage}{.45\linewidth}\centering\includegraphics[width=1\linewidth]{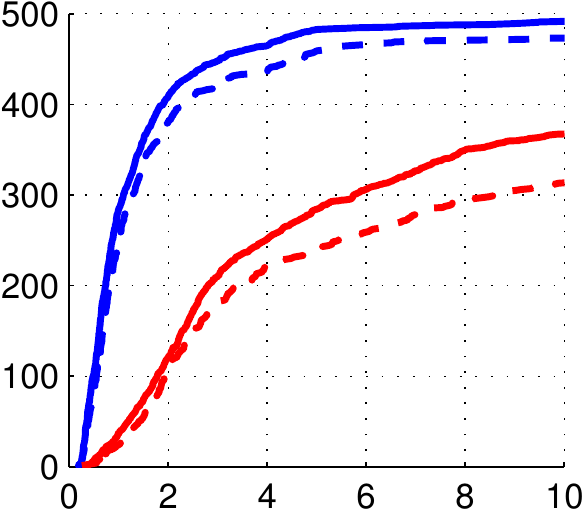}\\[-.5mm]\footnotesize ~~$e_\text{align}$\end{minipage}
\begin{minipage}{.45\linewidth}\centering\includegraphics[width=1\linewidth]{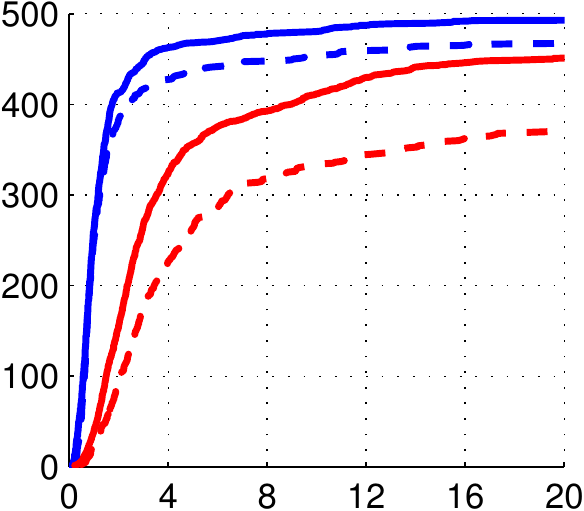}\\[-.5mm]\footnotesize ~~$e_r$ (degree)\end{minipage}\\
\begin{minipage}{0.03\linewidth}\rotatebox{90}{\footnotesize ~~~number of runs}\\\end{minipage}
\begin{minipage}{.91\linewidth}\includegraphics[width=0.95\linewidth]{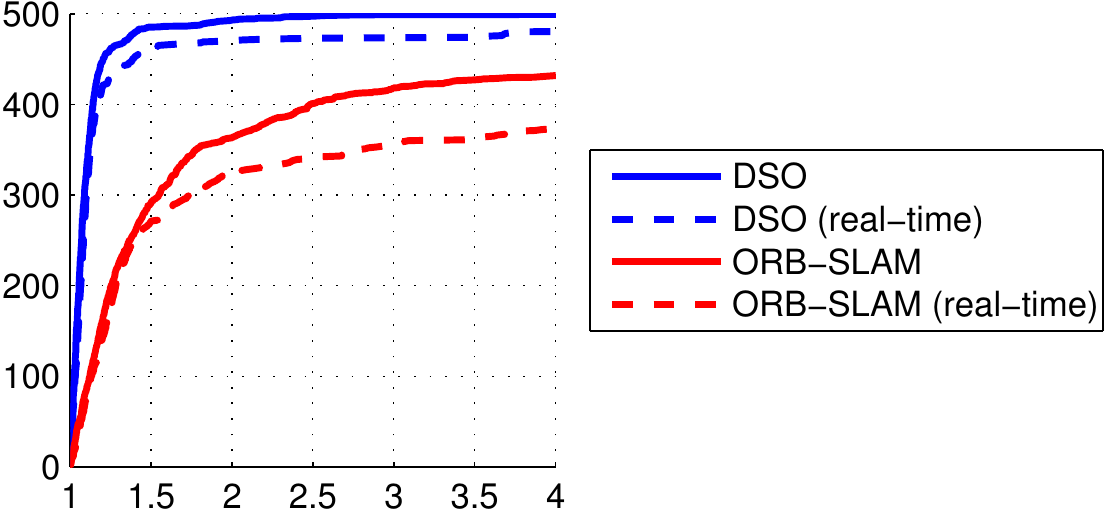}\\[-.5mm]\footnotesize {\color{white}\tiny secret Message} $e'_s$ (multiplier)\end{minipage}\\[2mm]
\caption{\textbf{Evaluation result.} Cumulative error-plots over 
all 50 sequences, run forwards and backwards, 5 times each to account for non-deterministic behaviour. For each 
error-value ($x$-axis), the plot shows the number of runs ($y$-axis) for which the achieved error was smaller. 
Note that since $e_s$ is a multiplicative factor, we summarize $e'_s = \max(e_s, e_s^{-1})$. The solid line
corresponds to non-real time execution, the dashed line to hard enforced real-time processing.}
\label{fig:acc}
\end{figure}

A common approach hence is to show only results on a hand-picked subset of sequences 
on which the compared methods do not fail (encouraging manual overfitting), or to show large tables with 
error values, which is not practicable for a dataset containing 50 sequences. 
A better approach is to summarize tracking accuracy as cumulative
distribution, visualizing on how many sequences the error is
below a certain threshold -- it shows both the accuracy on sequences
where a method works well, as well as the method's robustness, i.e.,
on how many sequences it does not fail.

Figure~\ref{fig:acc} shows such cumulative error-plots for two methods, 
DSO (Direct Sparse Odometry) \cite{engel16archiveOdometry} and ORB-SLAM \cite{mur2015orb}, evaluated on the presented dataset. 
Each of the 50 sequences is run 5 times forwards and 5 times backwards, giving a total of 500 runs for each line shown in the plots.

\begin{figure}
\centering
\includegraphics[width=.183\linewidth]{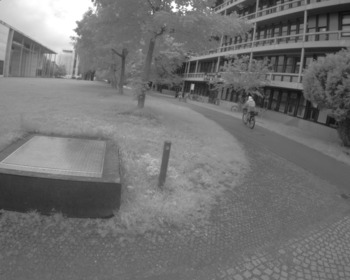}\hspace{-.5mm}
\includegraphics[width=.195\linewidth]{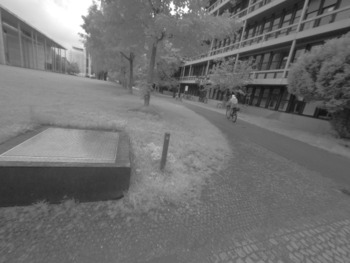}\hspace{-.5mm}
\includegraphics[width=.195\linewidth]{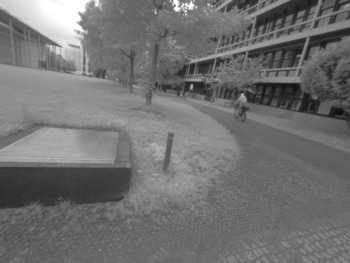}\hspace{-.5mm}
\includegraphics[width=.195\linewidth]{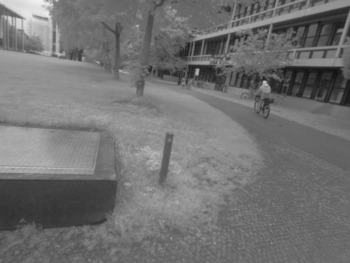}\hspace{-.5mm}
\includegraphics[width=.195\linewidth]{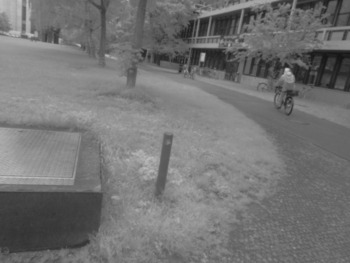}\\
\includegraphics[width=.183\linewidth]{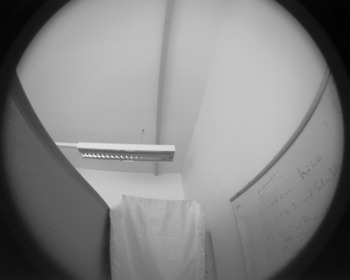}\hspace{-.5mm}
\includegraphics[width=.195\linewidth]{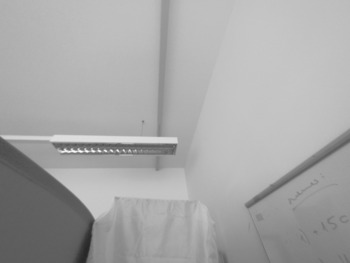}\hspace{-.5mm}
\includegraphics[width=.195\linewidth]{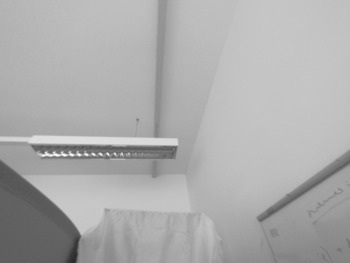}\hspace{-.5mm}
\includegraphics[width=.195\linewidth]{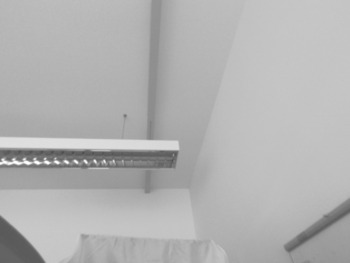}\hspace{-.5mm}
\includegraphics[width=.195\linewidth]{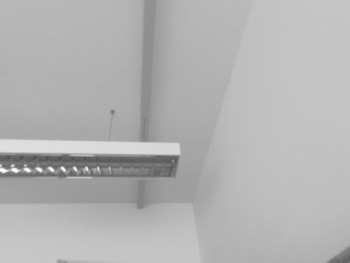}\\[-1.5mm]
{\hspace{5mm}\scriptsize original \hfill f=min \hfill f=300 \hfill f=400 \hfill f=500} ~~~~~~~~\\[2mm]
\begin{minipage}{0.03\linewidth}\rotatebox{90}{\footnotesize ~~~number of runs}\\\end{minipage}
\begin{minipage}{.95\linewidth}\centering\includegraphics[width=1\linewidth]{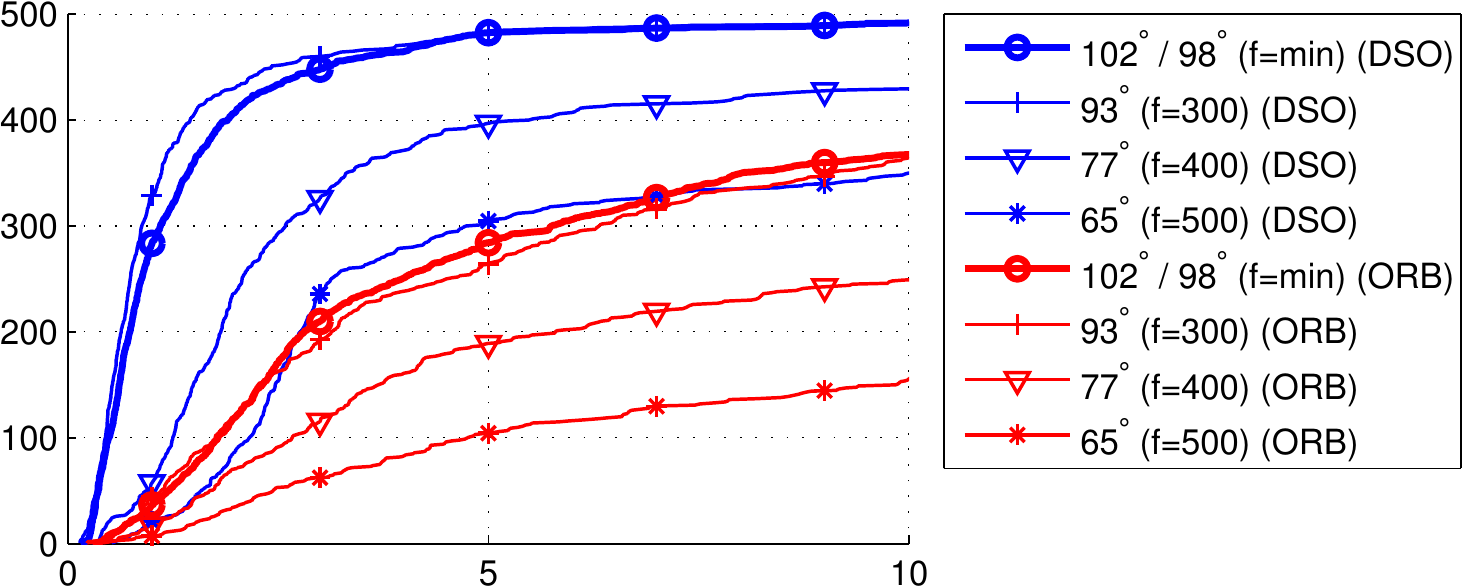}\\[-1mm]\footnotesize \hspace{-2.8cm} $e_\text{align}$\end{minipage}\\[1mm]
\caption{\textbf{Different field of view.} 
Alignment error when changing the horizontal field of view (i.e., using different focal lengths $f$ for the rectified images). 
The top shows two example images, rectified with the different focal lengths. \quotes{f=min} refers to the default setting, 
which differs for the two used lenses --
as comparison, the horizontal field of view of a Kinect camera is $63^\circ$. 
A smaller field of view significantly decreases accuracy and robustness, for both methods 
(note that for a cropped field of view, some sequences contain segments with only a white wall visible).}
\label{fig:FoV}
\end{figure}

\paragraph{Algorithm Parameter Settings. } Since both methods do not support the FOV camera model, we run the evaluation on 
pinhole-rectified images with VGA ($640 \times 480$) resolution.
Further, we disable explicit loop-closure detection and re-localization to allow application of our metric, 
and reduce the threshold where ORB-SLAM decides it is lost to 10 inlier observations. 
Note that we do not impose any restriction on implicit, \quotes{small} loop-closures, as long as these are 
found by ORB-SLAM's local mapping component (i.e., are included in the co-visibility graph).
Since DSO does not perform loop-closure or re-localization, we can use the default settings.
We run both algorithms in a non-real-time setting (at roughly one quarter speed), allowing to use
20 dedicated workstations with different CPUs to obtain the results presented in this paper. 
Figure~\ref{fig:acc} additionally shows results obtained when hard-enforcing real-time 
execution (dashed lines), obtained on the same workstation which is equipped with an i7-4910MQ CPU.

\begin{figure}
\centering
\begin{minipage}{0.03\linewidth}\rotatebox{90}{\footnotesize ~~~number of runs}\\\end{minipage}
\begin{minipage}{.95\linewidth}\centering\includegraphics[width=1\linewidth]{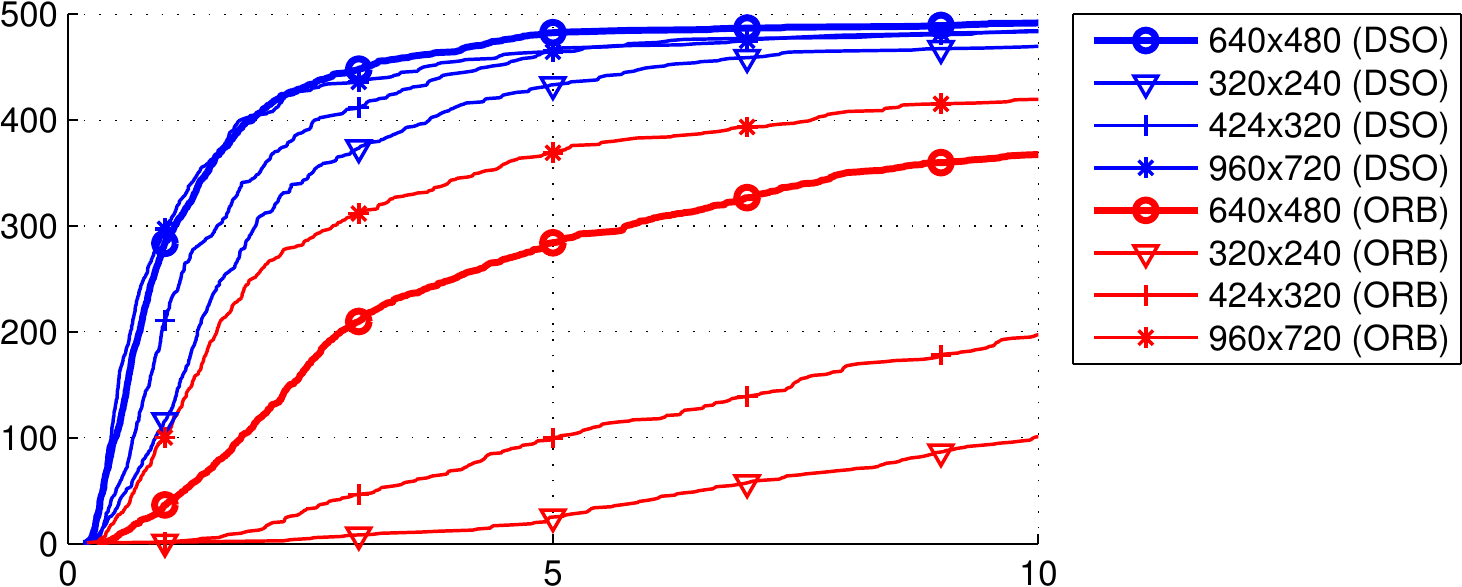}\\[-1mm]\footnotesize \hspace*{-2.2cm} $e_\text{align}$\end{minipage}\\[1mm]
\caption{\textbf{Different image resolution.} 
Alignment error when changing the rectified image resolution. Note that the algorithms were not
run in real time, hence the change in computational complexity is not accounted for.
While for ORB-SLAM the resolution has a strong effect, DSO is only marginally affected --
which is due to the sub-pixel-accurate nature of direct approaches.}
\label{fig:resolution}
\end{figure}

\begin{figure}
\centering
\begin{minipage}{0.03\linewidth}\rotatebox{90}{\footnotesize ~~~number of runs}\\\end{minipage}
\begin{minipage}{.95\linewidth}\centering\includegraphics[width=1\linewidth]{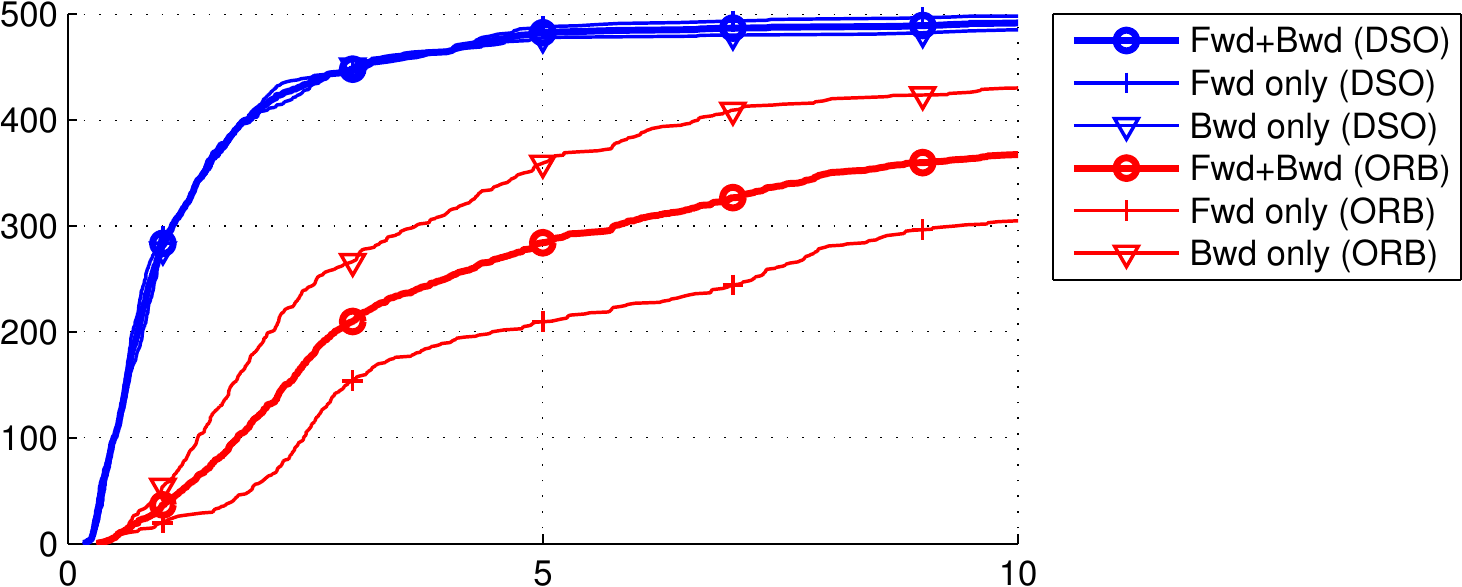}\\[-1mm]\footnotesize \hspace*{-2.3cm} $e_\text{align}$\end{minipage}\\[1mm]
\caption{\textbf{Dataset motion bias.}
Alignment error when running all sequences forwards and backwards, as well as the combination of both (default): While DSO is
largely unaffected by this, ORB-SLAM performs significantly better for backwards-motion. 
This is a classical example of \quotes{dataset bias}, and shows the importance of evaluating on large datasets,
covering a diverse range of environments and motion patterns.}
\label{fig:fwdbwd}
\end{figure}

\paragraph{Data Variations.} 
A good way to further to analyse the performance of an algorithm is to vary the sequences in a number of ways, 
simulating different real-world scenarios: 
\begin{itemize}
	 \item Figure \ref{fig:FoV} shows the tracking accuracy when rectifying the images to different 
		fields of view, while keeping the same resolution ($640 \times 480$). Since the raw data has a resolution of $1280\times 1024$, 
		the caused distortion is negligible.
	 \item Figure \ref{fig:resolution} shows the tracking accuracy when rectifying the images to 
	 	different resolutions, while keeping the same field of view. 
	 \item Figure \ref{fig:fwdbwd} shows the tracking accuracy when playing sequences only \textit{forwards} compared to the
	 results obtained when playing them only \textit{backwards} -- switching between predominantly forward-motion and predominantly backward-motion.
\end{itemize}
In each of the three figures, the bold lines correspond to the default parameter settings, which are the same across all evaluations.
We further give some example results and corresponding video snippets in Table~\ref{tab:examples}.

\begin{table*}
\begin{center}
\def\tabcolsep{0pt}
\begin{tabular}{ ccll } 
  \hspace{.5mm} {\footnotesize \textbf{Sequence}} \hspace{.5mm}& {\footnotesize \textbf{Selected Video Frames}} &  \hspace{6mm} {\footnotesize \hspace*{-1mm}\textbf{DSO}} & \hspace{0.2mm} {\footnotesize \hspace*{-2mm}\textbf{ORB-SLAM}} \\ 
 \raisebox{4mm}{\begin{minipage}{1cm}{\footnotesize \texttt{sq\_01}\\[-1mm]1:35 min\\[-1mm]50 fps}\end{minipage}}&
 \includegraphics[width=1.2cm]{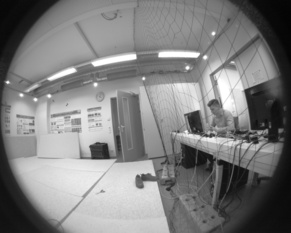}\hspace{-0.6mm}
 \includegraphics[width=1.2cm]{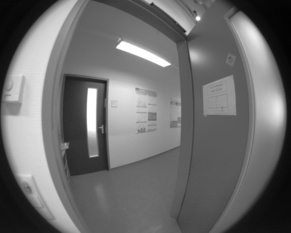}\hspace{-0.6mm}
 \includegraphics[width=1.2cm]{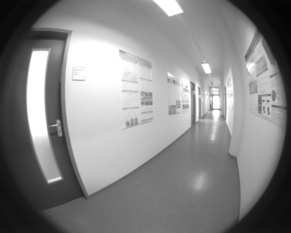}\hspace{-0.6mm}
 \includegraphics[width=1.2cm]{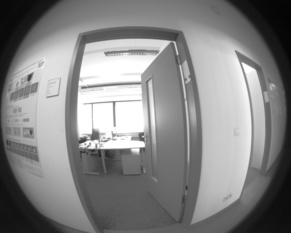}\hspace{-0.6mm}
 \includegraphics[width=1.2cm]{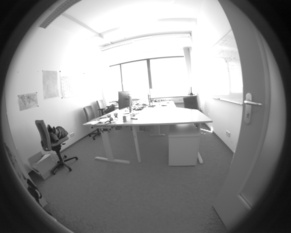}\hspace{-0.6mm}
 \includegraphics[width=1.2cm]{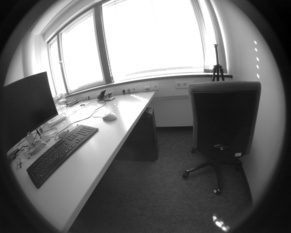}\hspace{-0.6mm}
 \includegraphics[width=1.2cm]{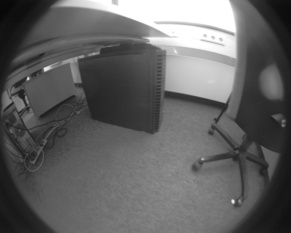}\hspace{-0.6mm}
 \includegraphics[width=1.2cm]{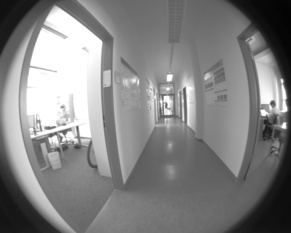}\hspace{-0.6mm}
 \includegraphics[width=1.2cm]{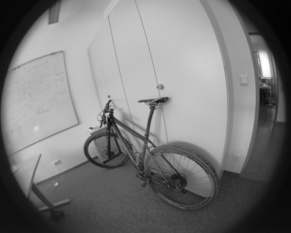}\hspace{-0.6mm}
 \includegraphics[width=1.2cm]{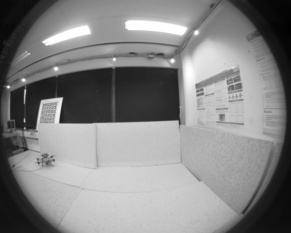}\hspace{-2mm}\vspace{-0.8mm}&  \hspace{2mm}
 \hspace*{-1mm}\raisebox{4mm}{
\begin{minipage}{1.8cm}{\tiny $e_\text{align}$ = 0.6 (0.5--0.8)\\[-1.5mm]$e_\text{r}$ = 0.61 (0.37--0.83)\\[-1.5mm]$e_\text{s}$ = 0.93 (0.91--0.94)}\end{minipage}
}&
 \hspace*{-2mm}\raisebox{4mm}{
\begin{minipage}{1.8cm}{\tiny $e_\text{align}$ = 2.8 (1.9--6.2)\\[-1.5mm]$e_\text{r}$ = 1.91 (1.18--9.72)\\[-1.5mm]$e_\text{s}$ = 0.70 (0.46--0.80)}\end{minipage}
} \\

 \raisebox{4mm}{\begin{minipage}{1cm}{\footnotesize \texttt{sq\_21}\\[-1mm]4:33 min\\[-1mm]20 fps}\end{minipage}}&	
 \includegraphics[width=1.2cm]{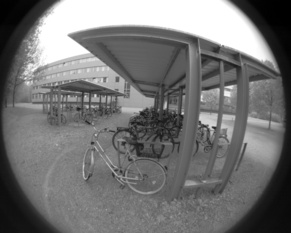}\hspace{-0.6mm}
 \includegraphics[width=1.2cm]{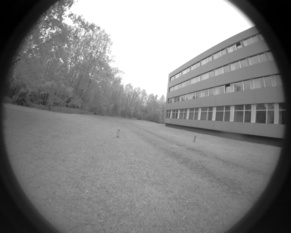}\hspace{-0.6mm}
 \includegraphics[width=1.2cm]{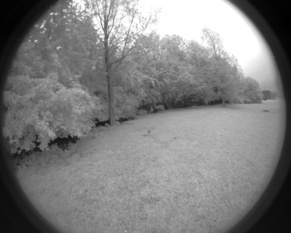}\hspace{-0.6mm}
 \includegraphics[width=1.2cm]{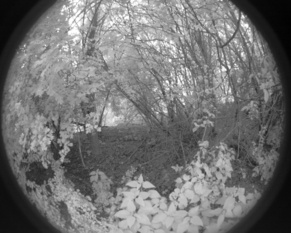}\hspace{-0.6mm}
 \includegraphics[width=1.2cm]{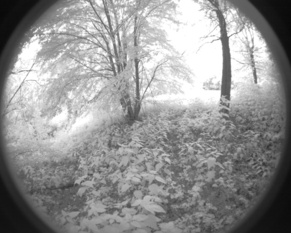}\hspace{-0.6mm}
 \includegraphics[width=1.2cm]{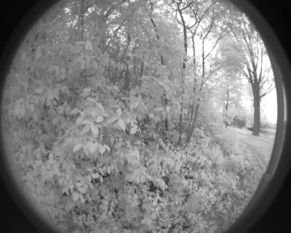}\hspace{-0.6mm}
 \includegraphics[width=1.2cm]{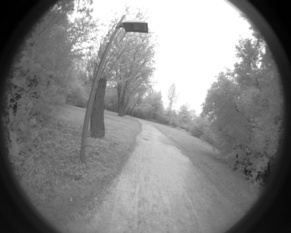}\hspace{-0.6mm}
 \includegraphics[width=1.2cm]{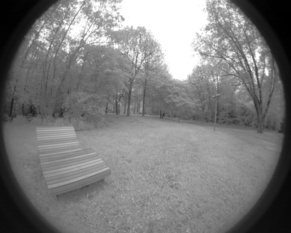}\hspace{-0.6mm}
 \includegraphics[width=1.2cm]{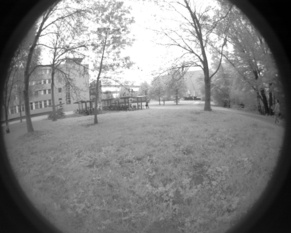}\hspace{-0.6mm}
 \includegraphics[width=1.2cm]{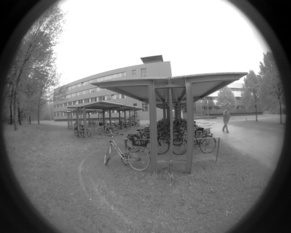}\hspace{-2mm}\vspace{-0.8mm}& \hspace{2mm}
 \hspace*{-1mm}\raisebox{4mm}{
\begin{minipage}{1.8cm}{\tiny $e_\text{align}$ = 4.2 (3.7--4.5)\\[-1.5mm]$e_\text{r}$ = 0.83 (0.76--0.98)\\[-1.5mm]$e_\text{s}$ = 0.74 (0.73--0.77)}\end{minipage}
}&
 \hspace*{-2mm}\raisebox{4mm}{
\begin{minipage}{1.8cm}{\tiny $e_\text{align}$ = 155 (127--254)\\[-1.5mm]$e_\text{r}$ = 6.0 (3.0--40.7)\\[-1.5mm]$e_\text{s}$ = 0.03 (0.01--0.04)}\end{minipage}
 } \\

  \raisebox{4mm}{\begin{minipage}{1cm}{\footnotesize \texttt{sq\_31}\\[-1mm]2:33 min\\[-1mm]21 fps}\end{minipage}}&
 \includegraphics[width=1.2cm]{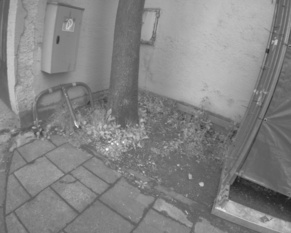}\hspace{-0.6mm}
 \includegraphics[width=1.2cm]{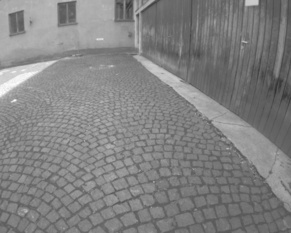}\hspace{-0.6mm}
 \includegraphics[width=1.2cm]{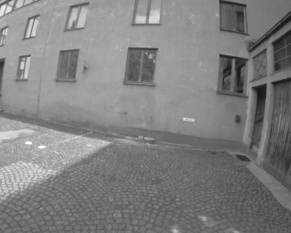}\hspace{-0.6mm}
 \includegraphics[width=1.2cm]{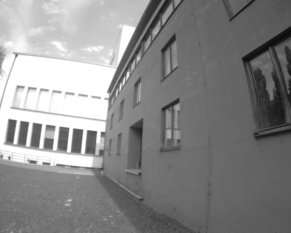}\hspace{-0.6mm}
 \includegraphics[width=1.2cm]{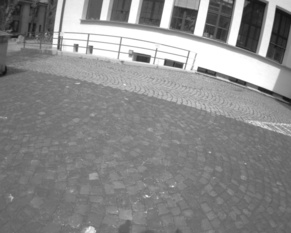}\hspace{-0.6mm}
 \includegraphics[width=1.2cm]{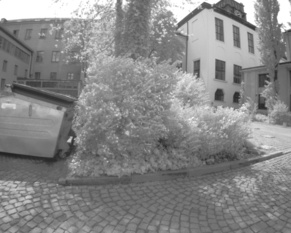}\hspace{-0.6mm}
 \includegraphics[width=1.2cm]{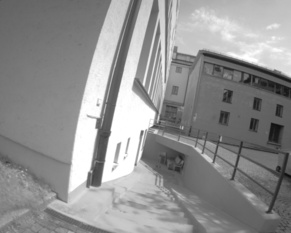}\hspace{-0.6mm}
 \includegraphics[width=1.2cm]{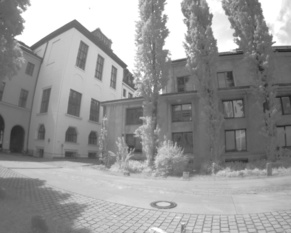}\hspace{-0.6mm}
 \includegraphics[width=1.2cm]{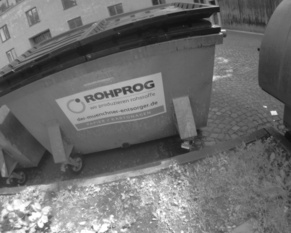}\hspace{-0.6mm}
 \includegraphics[width=1.2cm]{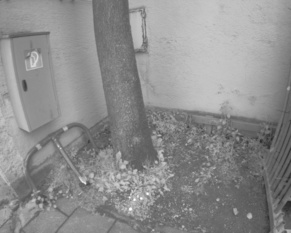}\hspace{-2mm}\vspace{-0.8mm}&  \hspace{2mm}
 \hspace*{-1mm}\raisebox{4mm}{
\begin{minipage}{1.8cm}{\tiny $e_\text{align}$ = 0.6 (0.6--0.7)\\[-1.5mm]$e_\text{r}$ = 1.34 (1.19--1.42)\\[-1.5mm]$e_\text{s}$ = 0.94 (0.93--0.97)}\end{minipage}
}&
 \hspace*{-2mm}\raisebox{4mm}{
\begin{minipage}{1.8cm}{\tiny $e_\text{align}$ = 5.7 (4.8--7.2)\\[-1.5mm]$e_\text{r}$ = 0.90 (0.59--3.68)\\[-1.5mm]$e_\text{s}$ = 0.57 (0.51--0.62)}\end{minipage}
} \\

  \raisebox{4mm}{\begin{minipage}{1cm}{\footnotesize \texttt{sq\_38}\\[-1mm]2:13 min\\[-1mm]25 fps}\end{minipage}}&
 \includegraphics[width=1.2cm]{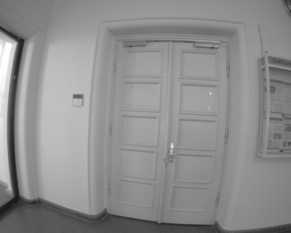}\hspace{-0.6mm}
 \includegraphics[width=1.2cm]{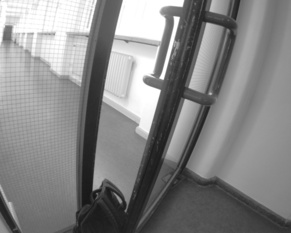}\hspace{-0.6mm}
 \includegraphics[width=1.2cm]{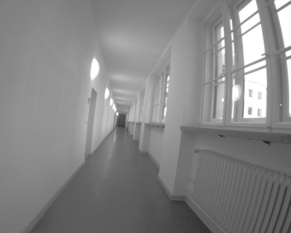}\hspace{-0.6mm}
 \includegraphics[width=1.2cm]{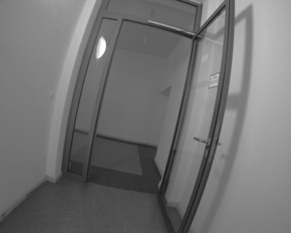}\hspace{-0.6mm}
 \includegraphics[width=1.2cm]{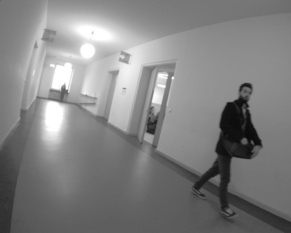}\hspace{-0.6mm}
 \includegraphics[width=1.2cm]{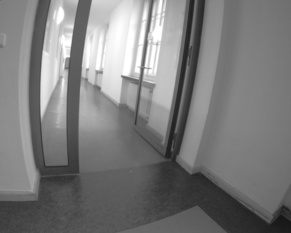}\hspace{-0.6mm}
 \includegraphics[width=1.2cm]{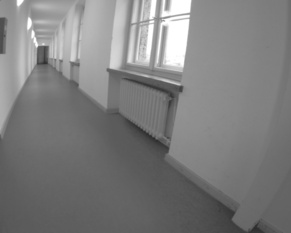}\hspace{-0.6mm}
 \includegraphics[width=1.2cm]{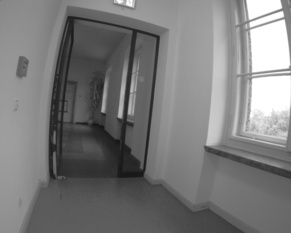}\hspace{-0.6mm}
 \includegraphics[width=1.2cm]{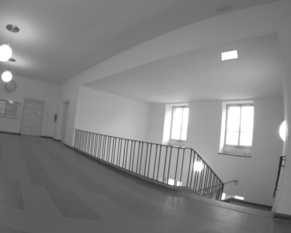}\hspace{-0.6mm}
 \includegraphics[width=1.2cm]{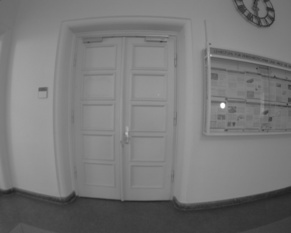}\hspace{-2mm}\vspace{-0.8mm}&  \hspace{2mm}
 \hspace*{-1mm}\raisebox{4mm}{
\begin{minipage}{1.8cm}{\tiny $e_\text{align}$ = 0.5 (0.5--0.6)\\[-1.5mm]$e_\text{r}$ = 1.06 (0.80--1.40)\\[-1.5mm]$e_\text{s}$ = 1.04 (1.02--1.04)}\end{minipage}
}&
 \hspace*{-2mm}\raisebox{4mm}{
\begin{minipage}{1.8cm}{\tiny $e_\text{align}$ = 28 (13--$\infty$)\\[-1.5mm]$e_\text{r}$ = 19.3 (4.6--$\infty$)\\[-1.5mm]$e_\text{s}$ = 0.34 (0.16--$\infty$)}\end{minipage}
} \\

  \raisebox{4mm}{\begin{minipage}{1cm}{\footnotesize \texttt{sq\_50}\\[-1mm]2:41 min\\[-1mm]25 fps}\end{minipage}}&
 \includegraphics[width=1.2cm]{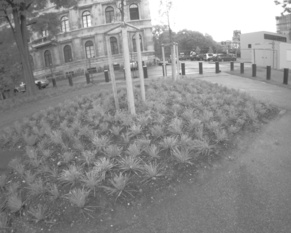}\hspace{-0.6mm}
 \includegraphics[width=1.2cm]{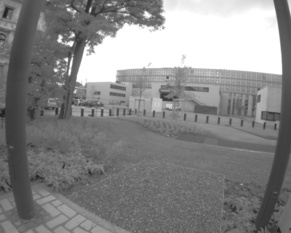}\hspace{-0.6mm}
 \includegraphics[width=1.2cm]{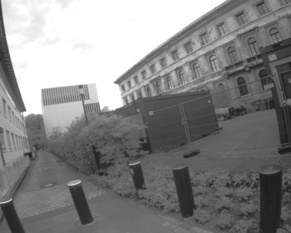}\hspace{-0.6mm}
 \includegraphics[width=1.2cm]{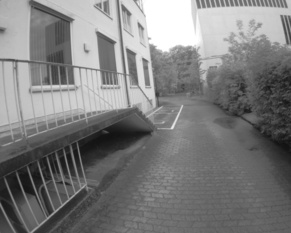}\hspace{-0.6mm}
 \includegraphics[width=1.2cm]{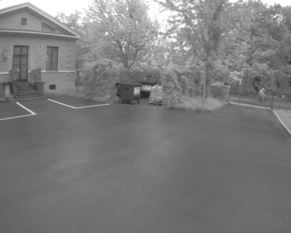}\hspace{-0.6mm}
 \includegraphics[width=1.2cm]{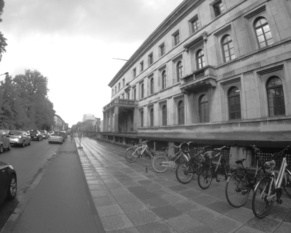}\hspace{-0.6mm}
 \includegraphics[width=1.2cm]{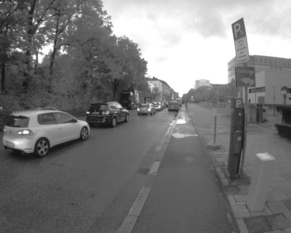}\hspace{-0.6mm}
 \includegraphics[width=1.2cm]{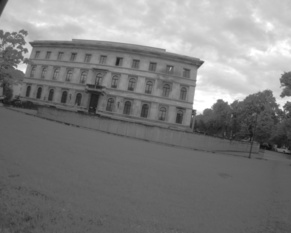}\hspace{-0.6mm}
 \includegraphics[width=1.2cm]{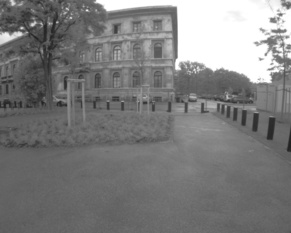}\hspace{-0.6mm}
 \includegraphics[width=1.2cm]{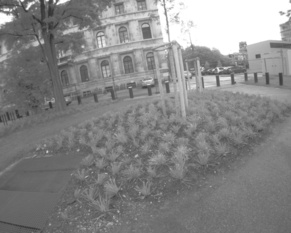}\hspace{-2mm}\vspace{-0.8mm}&  \hspace{2mm}
 \hspace*{-1mm}\raisebox{4mm}{
\begin{minipage}{1.8cm}{\tiny $e_\text{align}$ = 0.8 (0.8--1.0)\\[-1.5mm]$e_\text{r}$ = 0.62 (0.54--0.72)\\[-1.5mm]$e_\text{s}$ = 0.99 (0.97--1.01)}\end{minipage}
}&
 \hspace*{-2mm}\raisebox{4mm}{
\begin{minipage}{1.8cm}{\tiny $e_\text{align}$ = 7.5 (5.7--864)\\[-1.5mm]$e_\text{r}$ = 2.40 (1.01--79)\\[-1.5mm]$e_\text{s}$ = 0.58 (0.02--0.66)}\end{minipage}
} 
\end{tabular}\\[1mm]
\caption{\textbf{Example results.} Errors (alignment error $e_\text{align}$, scale-drift $e_\text{s}$ as multiplier, and rotation-drift $e_\text{r}$ in degree), as well as some selected frames for five example sequences when running them forwards. We show the median, the minimum, and maximum error respectively over 10 independent runs.}\vspace{-3mm}
\label{tab:examples}
\end{center}
\end{table*}

\begin{figure}
\centering
\begin{minipage}{0.03\linewidth}\rotatebox{90}{\footnotesize ~~~number of runs}\\\end{minipage}
\begin{minipage}{.45\linewidth}\centering\includegraphics[width=1\linewidth]{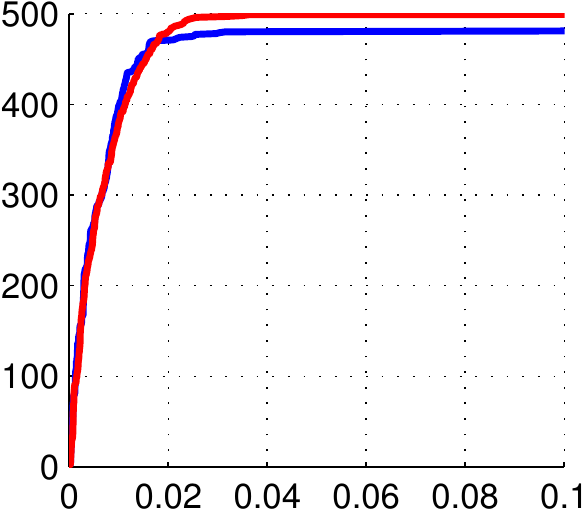}\\[-0mm]\footnotesize Start-segment RMSE\end{minipage}
\begin{minipage}{.45\linewidth}\centering\includegraphics[width=1\linewidth]{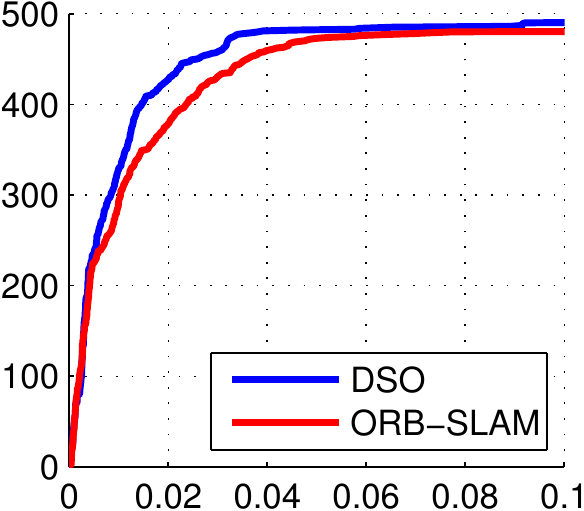}\\[-0mm]\footnotesize End-segment RMSE\end{minipage}\\[2mm]
\caption{\textbf{Start- and end-segment error.}
RMSE for alignment with the provided start- and end-segment ground truth. 
Note that it is roughly 100 times lower than $e_\text{align}$, and very similar for both evaluated methods. 
It can also be observed, that the automatic initialization of DSO fails occasionally 
(in that case all errors, including the start-segment error, are set to infinity), while this is not 
the case for ORB-SLAM. If the algorithm fails to provide 
an estimate for the full trajectory, the end-segment error is set to infinity.}
\label{fig:startEnd}
\end{figure}

\paragraph{Ground Truth Validation.} Since for most sequences, the used loop-closure ground truth is computed 
with a SLAM-algorithm (LSD-SLAM) itself, it is not perfectly accurate. We can however validate it by looking at the RMSE
when aligning the start- and end-segment, i.e., the minima of (\ref{eq:AlignS}) and (\ref{eq:AlignE}) respectively. 
They are summarized in Figure \ref{fig:startEnd}. Note the difference in order of magnitude: The RMSE within start- and end-segment is 
roughly 100 times smaller than the alignment RMSE, and very similar for both evaluated methods. This is a
strong indicator that almost all of the alignment error originates from accumulated drift, and not from noise in the ground truth.

\subsection{Dataset}
The full dataset, as well as preview-videos for all sequences are available on
\begin{center}
	\url{http://vision.in.tum.de/mono-dataset}
\end{center}
We provide
\begin{itemize}\itemsep-.3mm
	 \item Raw camera images of all 50 sequences (43GB; 190'000 frames in total), with frame-wise exposure times and computed 
	 ground truth alignment of start- and end-segment.
	 \item Geometric (FOV distortion model) and photometric calibrations (vignetting and response function). 
	 \item Calibration datasets (13GB) containing (1) checkerboard-images for geometric calibration, (2) several sequences
	 suitable for the proposed vignette and response function calibration, (3) images of a uniformly lit white paper.
	 \item Minimal c++ code for reading, pinhole-rectifying, and photometrically undistorting the images,
	 as well as for performing photometric calibration as proposed in Section~\ref{ssPhotoCalib}.
	 \item Matlab scripts to compute the proposed error metrics, as well as the raw tracking 
	 data for all runs used to create the plots and figures in Section~\ref{sec:Benchmark}.
\end{itemize}

\subsection{Known Issues}
\begin{itemize}
	\item Even though we use industry-grade cameras, the SDK provided by the manufacturer only allows to 
	asynchronously query the current exposure time. Thus, in some rare cases, the logged exposure 
	time may be shifted by one frame.
\end{itemize}

{\small
\bibliographystyle{ieee}
\bibliography{main}
}

\end{document}